\documentclass{article}

\PassOptionsToPackage{numbers, compress}{natbib}
\usepackage[preprint]{neurips_2026}

\usepackage[utf8]{inputenc} 
\usepackage[T1]{fontenc}    
\usepackage{hyperref}       
\usepackage{url}            
\usepackage{graphicx}        
\usepackage{booktabs}       
\usepackage{amsmath}        
\usepackage{amsfonts}       
\usepackage{nicefrac}       
\usepackage{microtype}      
\usepackage[table]{xcolor}      
\usepackage{multirow}          
\usepackage{tcolorbox}         
\tcbuselibrary{breakable, skins}
\usepackage{listings}          
\usepackage[varqu,varl]{inconsolata} 
\usepackage{fontawesome5}      
\usepackage{enumitem}          
\usepackage{pifont}            
\usepackage{float}             
\usepackage{placeins}          

\definecolor{prompt-bg}{RGB}{246,250,254}        
\definecolor{prompt-border}{RGB}{150,180,210}    
\definecolor{prompt-title-bg}{RGB}{70,105,150}   
\definecolor{prompt-title-fg}{RGB}{255,255,255}  
\definecolor{prompt-leftbar}{RGB}{110,145,185}   

\lstdefinestyle{promptstyle}{
    basicstyle=\ttfamily\footnotesize,
    breaklines=true,
    breakindent=0pt,
    breakatwhitespace=true,
    backgroundcolor=\color{prompt-bg},
    frame=none, numbers=none,
    columns=fullflexible,
    keepspaces=true,
    showspaces=false,
    showstringspaces=false,
    tabsize=2,
    upquote=true,
    escapeinside={<@}{@>},
    aboveskip=2pt, belowskip=2pt,
    moredelim=[is][\bfseries]{**}{**},
}

\newtcolorbox{promptbox}[1]{
    enhanced, unbreakable,
    arc=2pt, boxrule=0.6pt,
    colframe=prompt-leftbar!50!prompt-border,
    colback=prompt-bg,
    coltitle=prompt-title-fg,
    fonttitle=\bfseries\small,
    title={\strut #1},
    titlerule=0pt,
    toptitle=3pt, bottomtitle=3pt,
    top=4pt, bottom=4pt, left=6pt, right=5pt,
    boxsep=0pt,
    colbacktitle=prompt-title-bg,
    before title={\faCode\hspace{0.25em}},
}

\title{Think When Needed: Adaptive Reasoning-Driven Multimodal Embeddings with a Dual-LoRA Architecture}

\renewcommand{\thefootnote}{\fnsymbol{footnote}}
\author{%
  Longxiang Zhang\textsuperscript{*} \quad
  Weilong Dai\textsuperscript{*\;\dag} \quad
  Guanghao Zhang \quad
  Hao Jiang\textsuperscript{\ddag} \quad
  Pipei Huang \\
  Alibaba Group \\
  \texttt{\{shengxiang.zlx, junmu.dwl, guanghao.zgh, aoshu.jh, pipei.hpp\}@taobao.com}
}

\begin{document}
\raggedbottom

\maketitle
\renewcommand{\thefootnote}{\arabic{footnote}}
{\renewcommand{\thefootnote}{}\footnotetext{\textsuperscript{*}\,Equal contribution. \quad \textsuperscript{\dag}\,Project lead. \quad \textsuperscript{\ddag}\,Corresponding author.}\addtocounter{footnote}{-1}}

\begin{abstract}
	Multimodal large language models (MLLMs) have emerged as a powerful backbone for multimodal embeddings.
	Recent methods introduce chain-of-thought (CoT) reasoning into the embedding pipeline to improve retrieval quality, but remain costly in both model size and inference cost. They typically employ separate reasoner and embedder with substantial parameter overhead, and generate CoT indiscriminately for every input. However, we observe that for simple inputs, discriminative embeddings already perform well, and redundant reasoning can even mislead the model, degrading performance.
	To address these limitations, we propose \textbf{T}hink \textbf{W}hen \textbf{N}eeded (TWN), a unified multimodal embedding framework with adaptive reasoning.
	TWN introduces a dual-LoRA architecture that attaches reasoning and embedding adapters to a shared frozen backbone, detaching gradients at their interface to mitigate gradient conflicts introduced by joint optimization while keeping parameters close to a single model.
	Building on this, an adaptive think mechanism uses a self-supervised routing gate to decide per input whether to generate CoT, skipping unnecessary reasoning to reduce inference overhead and even improve retrieval quality.
	We further explore embedding-guided RL to optimize CoT quality beyond supervised training.
	On the 78 tasks of MMEB-V2, TWN achieves state-of-the-art embedding quality while being substantially more efficient than existing generative methods, requiring only 3--5\% additional parameters relative to the backbone and up to 50\% fewer reasoning tokens compared to the full generative mode.
	Our project page is at \url{https://github.com/winterfell00/Think-When-Needed}.
\end{abstract}


\section{Introduction}

\begin{figure*}[t]
	\centering
	\includegraphics[width=\textwidth]{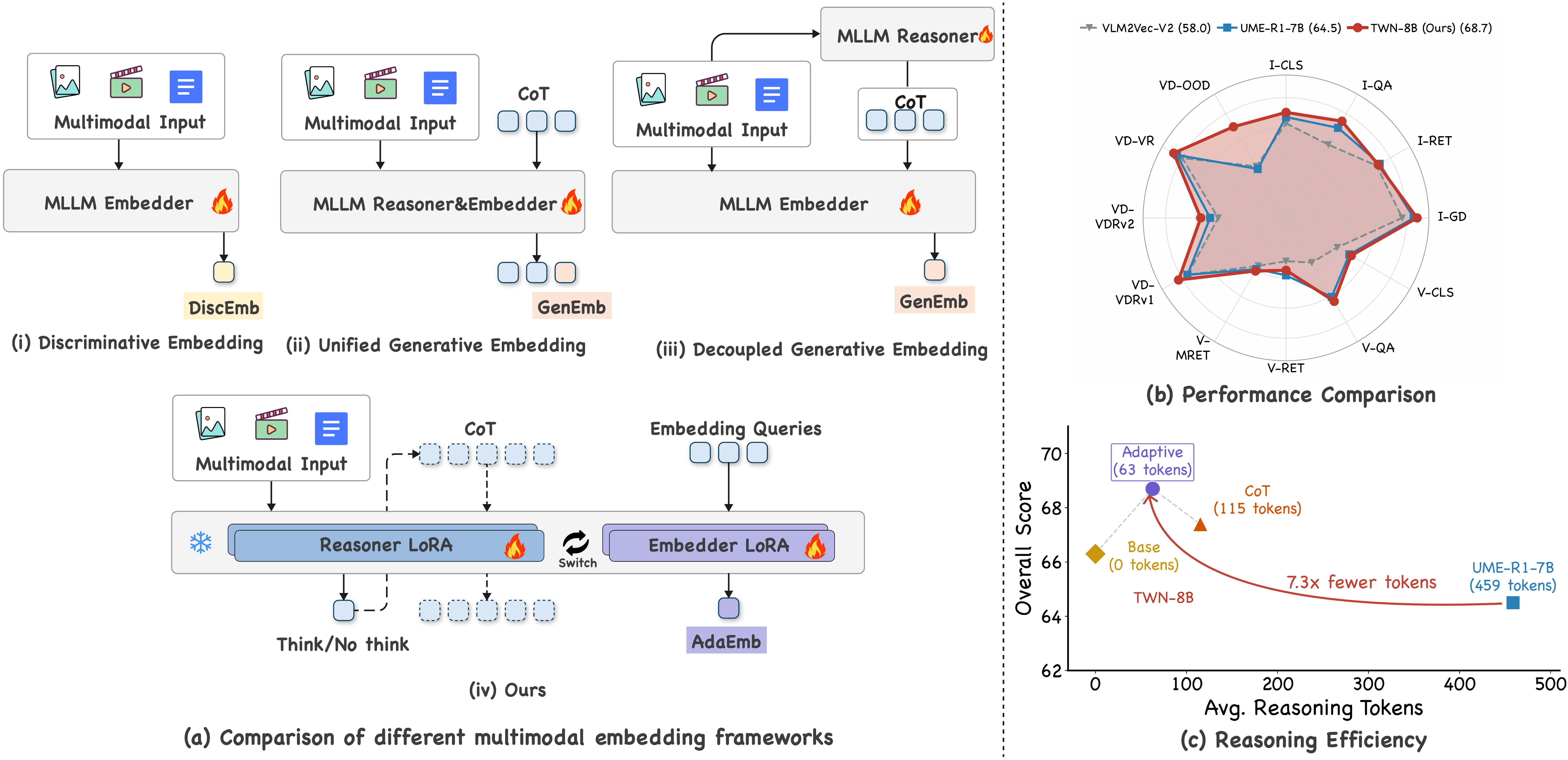}
	\caption{(a)~Comparison of multimodal embedding frameworks: (i)~discriminative, (ii)~unified generative, (iii)~decoupled generative, and (iv)~TWN (ours). (b)~Performance on MMEB-V2. (c)~Average reasoning tokens per input.}
	\label{fig:framework}
\end{figure*}

Multimodal embeddings map heterogeneous inputs such as images, videos, visual documents, and text into a unified representation space, serving as a foundational component for cross-modal retrieval, multimodal retrieval-augmented generation~\cite{lewis2020retrieval}, and related applications.
Pioneering studies adopted a CLIP-like dual-encoder architecture~\cite{radford2021learning,jia2021align,zhai2023sigmoid} and learned aligned cross-modal representations through large-scale image-text contrastive pre-training. However, such architectures struggle to handle interleaved multimodal inputs or to support complex, instruction-aware retrieval.
With the rapid progress of multimodal large language models (MLLMs)~\cite{liu2023llava,li2023blip2,chen2024internvl,wang2024qwen2vl}, the field has shifted toward adopting MLLMs as unified encoders~\cite{jiang2024e5v,jiang2024vlm2vec,lin2024mmembed,lan2025llave,gu2025unime}, which natively handle interleaved multimodal inputs and enable more complex retrieval scenarios through their strong understanding capabilities.

Prevailing MLLM-based embedding methods feed the input through an MLLM and take the last-layer hidden state of the final token as the semantic representation, which we refer to as \emph{discriminative embeddings} (Figure~\ref{fig:framework}(a-i)).
This line of work treats the MLLM as a generic feature extractor and does not explicitly leverage its generation and reasoning abilities, and therefore tends to fall short for hard queries that demand fine-grained understanding and reasoning.
Consequently, recent studies explicitly incorporate the reasoning ability of MLLMs into the embedding pipeline, yielding what we call \emph{generative embeddings}~\cite{cui2025think,lan2025ume,liu2025rge,liu2025lamra}.
Concretely, a \emph{reasoner} first produces a chain-of-thought (CoT)~\cite{wei2022chain} for the input, and an \emph{embedder} then derives the semantic representation conditioned on both the original input and the generated CoT.
These methods achieve noticeably better retrieval quality on hard queries than their discriminative counterparts.

Despite this progress, current generative embedding methods remain limited in two key respects.
\emph{Architecturally}, decoupled methods (Figure~\ref{fig:framework}(a-iii); e.g., TTE~\cite{cui2025think}) employ separate models for reasoning and embedding, incurring substantial parameter overhead.
Unified methods (Figure~\ref{fig:framework}(a-ii); e.g., UME-R1~\cite{lan2025ume}) share a single MLLM for both tasks; while parameter-efficient, the competing autoregressive and contrastive objectives may introduce gradient conflicts~\cite{yu2020gradient} that limit performance.
In terms of inference, all existing methods generate CoT indiscriminately for every input, incurring substantial decoding overhead; yet for simple inputs, discriminative embeddings already perform well~\cite{lan2025ume}, and unnecessary CoT can actively mislead the model, degrading retrieval quality.

To address these limitations, we propose \textbf{T}hink \textbf{W}hen \textbf{N}eeded (TWN), a unified multimodal embedding framework with adaptive reasoning (Figure~\ref{fig:framework}(a-iv)).
Our key ideas are: (1)~a \textbf{dual-LoRA} architecture that attaches separate reasoning and embedding adapters to a shared frozen backbone, mitigating gradient conflicts while keeping parameters close to a single model; and (2)~an \textbf{adaptive think} mechanism that selectively invokes CoT only when it benefits retrieval, skipping unnecessary reasoning for simple inputs.
Our contributions are:
\begin{itemize}[nosep,leftmargin=1.2em]
	\item We propose a \textbf{dual-LoRA architecture} that attaches reasoning and embedding adapters to a shared frozen backbone with gradient detachment at their interface, combining the gradient isolation of decoupled methods with the parameter efficiency of unified methods while adding only a small fraction of backbone parameters.
	\item We introduce an \textbf{adaptive think mechanism} with a routing gate that decides per input whether to generate CoT, skipping unnecessary reasoning to reduce inference cost and avoiding misleading CoT that can degrade retrieval quality on simple inputs.
	\item We explore \textbf{embedding-guided RL} that exploits the parameter separation of dual-LoRA to freeze the embedding adapter as a stationary reward environment and introduces a global embedding cache for more discriminative reward signals.
	\item On the 78 tasks of MMEB-V2, TWN achieves \textbf{state-of-the-art} embedding quality while being substantially more efficient, requiring only 3--5\% additional parameters relative to the backbone and up to 50\% fewer reasoning tokens compared to the full generative mode.
\end{itemize}


\section{Method}

\subsection{Data Construction}
\label{sec:data}

To provide high-quality CoT supervision for the reasoning adapter, we construct training data at scale through a generate-then-filter pipeline that spans diverse modalities (image, video, and visual document) and multiple datasets, and enforces alignment between reasoning traces and retrieval objectives through rigorous quality filtering.

We first curate source data from multimodal tasks, following the data paradigm of VLM2Vec-V2~\cite{meng2025vlm2vec}:
(1) image-based tasks from MMEB-train~\cite{jiang2024vlm2vec}, covering classification, question answering, retrieval, and grounding;
(2) video-language instruction data from LLaVA-Hound~\cite{zhang2024direct}, including captioning, question answering, and retrieval;
(3) visual document retrieval data from ViDoRe~\cite{faysse2024colpali} and VisRAG~\cite{yu2024visrag}.

For each retrieval pair $(q, t^+)$, we generate side-specific CoT traces independently for $q$ and $t^+$.
To prevent label leakage, the teacher model receives only one side at a time without access to its paired counterpart, producing structured traces in the format \texttt{<think>}\,\ldots\,\texttt{</think>}\texttt{<answer>}\,\ldots\,\texttt{</answer>},
where \texttt{<think>} captures step-by-step analysis and \texttt{<answer>} provides a retrieval-oriented summary.
The generation is guided by task-specific prompts, e.g., producing the most specific category label for classification and extracting key semantic elements for retrieval.

We then filter the generated traces using a judge model with task-adaptive validation modes: strict verification (reasoning quality and answer matching) for tasks with well-defined answers, and hallucination-only verification for open-ended tasks.
Samples that fail their respective criteria are removed so the model is trained only on CoT data that passes the filtering criteria.
The final dataset comprises retrieval pairs $(q_n, t_n^+)$ with side-specific CoT traces $c_n$. For the teacher and judge model, we use Qwen3.5-35B-A3B\footnote{\url{https://huggingface.co/Qwen/Qwen3.5-35B-A3B}}. More details are provided in Appendix~\ref{sec:dataset_construction}.

\subsection{Architecture}
\label{sec:architecture}

\begin{figure*}[t]
	\centering
	\includegraphics[width=\textwidth]{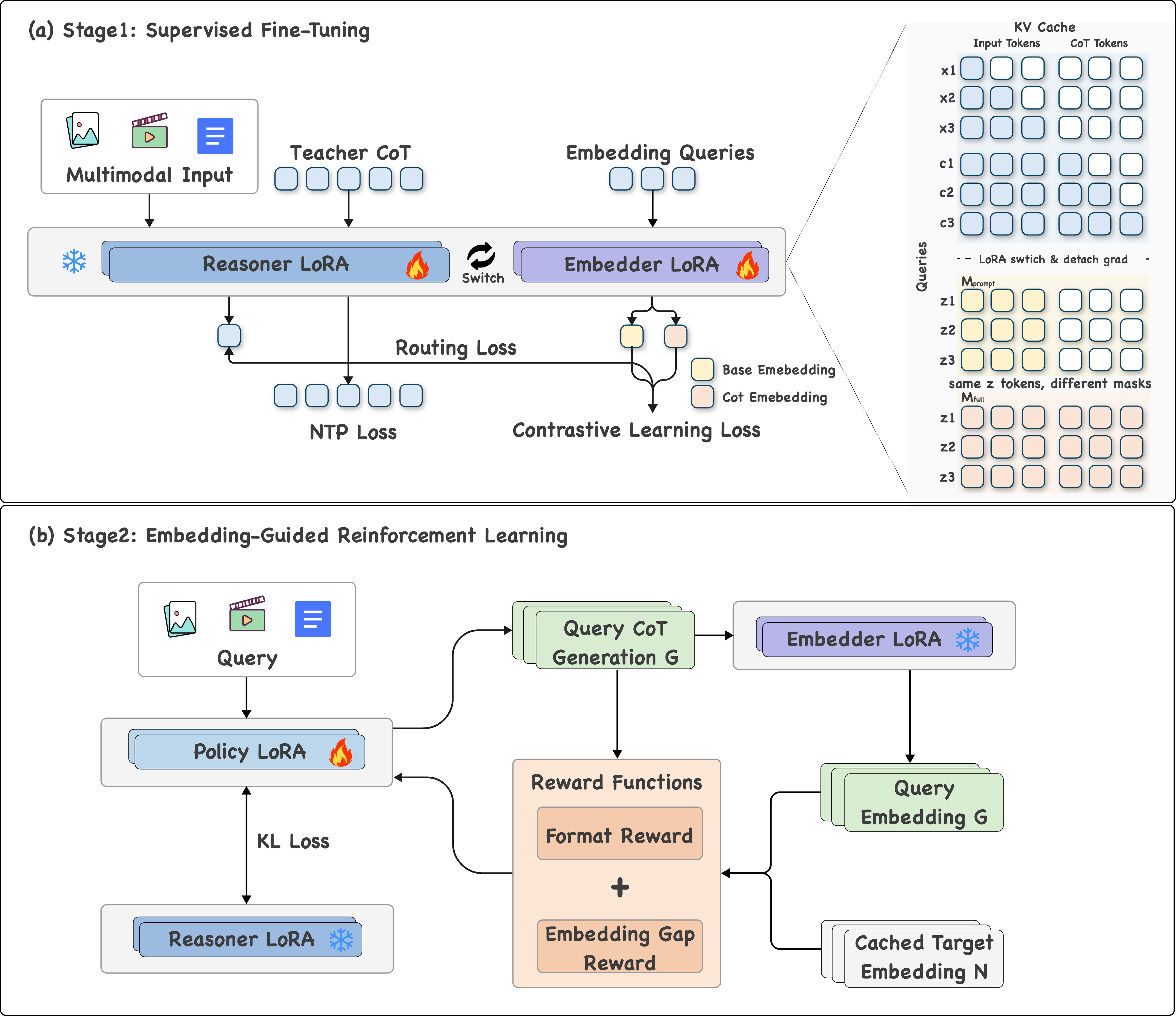}
	\caption{Overview of TWN\@. (a)~Stage~1: Supervised fine-tuning jointly trains the reasoning and embedding LoRA adapters with NTP, contrastive, and routing losses. The right panel illustrates dual-mask embedding extraction from the shared KV cache. (b)~Stage~2: Embedding-guided reinforcement learning optimizes the reasoning policy with the embedding adapter frozen as a stationary reward environment.}
	\label{fig:method}
\end{figure*}

TWN comprises two core components: a \emph{dual-LoRA} structure that isolates the gradient flows of reasoning and embedding within a single backbone, and an \emph{adaptive think} mechanism that selectively triggers chain-of-thought per input.

\paragraph{Dual-LoRA.}
We attach two lightweight LoRA~\cite{hu2022lora} adapters to the same frozen MLLM backbone (Figure~\ref{fig:method}): a \emph{reasoning adapter} $\theta_r$ and an \emph{embedding adapter} $\theta_e$.
The reasoning adapter processes the input $x$ and optionally generates CoT; the embedding adapter then directly reuses the resulting KV cache (with gradients detached) and appends $K$ learnable query tokens $\mathbf{z} \in \mathbb{R}^{K \times d}$ to extract the final embedding without re-encoding the input.
Gradient detachment prevents backward gradient flow between the generative and discriminative objectives, mitigating gradient conflicts during joint training.

\paragraph{Adaptive Think.}
A \emph{routing gate} $g_\phi$ decides per input whether to generate CoT.
Since this decision must be made before CoT generation, we pass the last input token's hidden state $\mathbf{h}_p$ (produced by $\theta_r$) through a lightweight MLP with sigmoid activation $\sigma$:
\begin{equation}
	w = \sigma\!\bigl(\operatorname{MLP}(\mathbf{h}_p)\bigr) \in [0, 1].
	\label{eq:gate}
\end{equation}
At inference, $w$ is thresholded at $0.5$ to produce a binary decision: when $w < 0.5$, the model directly produces a \emph{base embedding} $\mathbf{h}^{\text{base}}$ from the input tokens alone; when $w \geq 0.5$, it first generates CoT and then derives a \emph{CoT-enhanced embedding} $\mathbf{h}^{\text{cot}}$ from both input and reasoning tokens.

Manually labeling per-input reasoning necessity is impractical.
We instead observe that the contrastive training process itself provides a natural self-supervised signal: within each batch, we can directly compare the retrieval quality of $\mathbf{h}^{\text{base}}$ and $\mathbf{h}^{\text{cot}}$.
If the CoT-enhanced embedding achieves a larger positive margin than the base embedding, reasoning is beneficial for that input, and the gate should activate.

Specifically, for the $i$-th input in the batch, we compute the positive margin:
\begin{equation}
	m(\mathbf{h}) = \cos(\mathbf{h}, \mathbf{h}^+) - \max_{j \neq i} \cos(\mathbf{h}, \mathbf{h}_j)
	\label{eq:margin}
\end{equation}
for both the base and CoT-enhanced embeddings, where $\mathbf{h}^+$ is the matched positive embedding and $\mathbf{h}_j$ are in-batch negatives, and derive a soft routing target:
\begin{equation}
	\hat{w} = \sigma\!\left(\frac{m(\mathbf{h}^{\text{cot}}) - m(\mathbf{h}^{\text{base}}) - \delta}{\tau_g}\right),
	\label{eq:routing_target}
\end{equation}
where $\delta$ is a margin offset and $\tau_g$ is a temperature that controls the sharpness of the target.
The routing loss minimizes the binary cross-entropy between the gate output $w$ and the routing target:
\begin{equation}
	\mathcal{L}_{\text{route}} = \operatorname{BCE}(w,\, \hat{w}).
	\label{eq:routing_loss}
\end{equation}

\subsection{Training Pipeline}
\label{sec:training}

We train TWN in two stages (Figure~\ref{fig:method}).
Stage~1 (supervised fine-tuning) jointly trains the reasoning and embedding adapters, teaching the model to follow the structured reasoning format and to extract effective representations.
Stage~2 (embedding-guided reinforcement learning) further optimizes the reasoning adapter using retrieval quality as the reward signal, enabling the model to generate CoT that is more aligned with retrieval objectives.

\subsubsection{Stage 1: Supervised Fine-Tuning}
\label{sec:sft}

\paragraph{Reasoning Objective.}
The reasoning adapter $\theta_r$ is trained with next-token prediction on the annotated CoT traces:
\begin{equation}
	\mathcal{L}_{\text{NTP}} = -\frac{1}{\sum_n T_n} \sum_{n=1}^{N} \sum_{t=1}^{T_n} \log p_{\theta_r}(c_{n,t} \mid x_n, c_{n,<t}),
	\label{eq:ntp_loss}
\end{equation}
where $x_n$ is the input, $c_n = (c_{n,1}, \dots, c_{n,T_n})$ is the annotated CoT trace of length $T_n$, and $N$ is the number of training samples.

\paragraph{Embedding Objective.}
The embedding adapter $\theta_e$ is trained with the InfoNCE~\cite{oord2018representation} contrastive loss:
\begin{equation}
	\mathcal{L}_{\text{CL}}(\mathbf{h}) = -\frac{1}{B} \sum_{i=1}^{B} \log \frac{\exp\bigl(\cos(\mathbf{h}_{q_i}, \mathbf{h}_{t_i}) / \tau\bigr)}{\sum_{j=1}^{B} \exp\bigl(\cos(\mathbf{h}_{q_i}, \mathbf{h}_{t_j}) / \tau\bigr)},
	\label{eq:infonce}
\end{equation}
where $B$ is the batch size and $\tau$ is the temperature.
We apply this loss to both embedding variants:
$\mathcal{L}_{\text{base}} = \mathcal{L}_{\text{CL}}(\mathbf{h}^{\text{base}})$ ensures the base embedding is effective independently;
$\mathcal{L}_{\text{cot}} = \mathcal{L}_{\text{CL}}(\mathbf{h}^{\text{cot}})$ directly optimizes the CoT-enhanced embedding.
During training, CoT traces are always present (teacher-annotated), so the KV cache contains both input and reasoning tokens.
To simultaneously obtain $\mathbf{h}^{\text{base}}$ and $\mathbf{h}^{\text{cot}}$ for contrastive and routing supervision, we extract them from the same KV cache using two attention masks.
The base embedding attends only to the input portion (via mask $\mathbf{M}_{\text{prompt}}$), while the CoT-enhanced embedding attends to the full sequence including reasoning tokens (via mask $\mathbf{M}_{\text{full}}$):
\begin{equation}
	\mathbf{h}^{\text{base}} = \operatorname{Normalize}\!\bigl(\operatorname{MeanPool}(f_{\theta_e}(\mathbf{z} \mid \mathrm{KV},\, \mathbf{M}_{\text{prompt}}))\bigr),
	\label{eq:base_emb}
\end{equation}
\begin{equation}
	\mathbf{h}^{\text{cot}} = \operatorname{Normalize}\!\bigl(\operatorname{MeanPool}(f_{\theta_e}(\mathbf{z} \mid \mathrm{KV},\, \mathbf{M}_{\text{full}}))\bigr),
	\label{eq:cot_emb}
\end{equation}
where $f_{\theta_e}$ denotes the forward pass of the embedding adapter, $\operatorname{MeanPool}$ averages over the $K$ query-token positions, and $\operatorname{Normalize}$ denotes $\ell_2$ normalization.

Together with the routing loss $\mathcal{L}_{\text{route}}$ (Equation~\ref{eq:routing_loss}), the total SFT loss is:
\begin{equation}
	\mathcal{L}_{\text{SFT}} = \mathcal{L}_{\text{NTP}} + \lambda_{\text{base}} \mathcal{L}_{\text{base}} + \lambda_{\text{cot}} \mathcal{L}_{\text{cot}} + \lambda_{\text{route}} \mathcal{L}_{\text{route}},
	\label{eq:sft_loss}
\end{equation}
where $\lambda_{\text{base}}$, $\lambda_{\text{cot}}$, and $\lambda_{\text{route}}$ are the loss weights for each objective.

\subsubsection{Stage 2: Embedding-Guided RL}

To further improve CoT quality beyond supervised learning, we use embedding-based reward signals to optimize the reasoning adapter via GRPO~\cite{shao2024deepseekmath}.

\paragraph{RL Configuration.}
We freeze all embedding components ($\theta_e$, learnable queries $\mathbf{z}$, routing gate $g_\phi$) as a stationary reward environment, and initialize the RL policy $\pi$ from the reasoning adapter $\theta_r$ as the sole trainable component.
The original $\theta_r$ remains frozen, inducing the reference policy $\pi_{\text{ref}}$ for KL regularization; since both are on the same backbone, computing $D_{\mathrm{KL}}(\pi \| \pi_{\text{ref}})$ requires only switching the active LoRA weights without additional model copies.

\paragraph{Reward Design.}
We define two reward signals for each candidate trace $c_i$.
The \emph{gap reward} measures how well the query embedding separates the positive target from negatives:
\begin{equation}
	R_{\text{gap}}(c_i) = \cos\!\bigl(\mathbf{h}_q(c_i), \mathbf{h}_{t^+}\bigr) - \mathbb{E}_{\tau_r}\!\bigl[\cos\!\bigl(\mathbf{h}_q(c_i), \mathbf{h}_{t^-}\bigr)\bigr],
	\label{eq:gap_reward}
\end{equation}
where $\mathbb{E}_{\tau_r}$ denotes a softmax-weighted expectation over negatives at temperature $\tau_r$ that smoothly up-weights hard negatives.
Since the limited pool of in-batch negatives yields noisy reward estimates, we pre-compute all target embeddings into a global cache $\mathcal{B} = \{\mathbf{h}_{t_j}\}_{j=1}^{|\mathcal{D}|}$ and sample negatives from $\mathcal{B}$ for more discriminative reward signals.
The \emph{format reward} encourages the model to follow the structured reasoning format:
\begin{equation}
	R_{\text{fmt}}(c_i) = \begin{cases} 1 & \text{if } c_i \text{ matches } \texttt{<think>}\,\ldots\,\texttt{</think>}\texttt{<answer>}\,\ldots\,\texttt{</answer>}, \\ 0 & \text{otherwise.} \end{cases}
	\label{eq:fmt_reward}
\end{equation}
The total reward is $R_i = R_{\text{gap}}(c_i) + R_{\text{fmt}}(c_i)$.

For each query, we sample $G$ candidate CoT traces from $\pi$ and pass each through the frozen embedding pipeline to obtain $\mathbf{h}_q(c_i)$.
The GRPO objective is:
\begin{equation}
	\mathcal{L}_{\text{GRPO}} = -\frac{1}{G} \sum_{i=1}^{G} \Bigl[\min\!\bigl(r_i \hat{A}_i,\; \operatorname{clip}(r_i, 1{-}\epsilon, 1{+}\epsilon)\, \hat{A}_i\bigr) - \beta\, D_{\mathrm{KL}}\!\bigl(\pi \| \pi_{\text{ref}}\bigr)\Bigr],
	\label{eq:grpo}
\end{equation}
where $r_i = \pi(c_i \mid q) / \pi_{\text{old}}(c_i \mid q)$ is the importance ratio with $\pi_{\text{old}}$ being the policy before the current update, $\hat{A}_i = (R_i - \mu_R) / \sigma_R$ is the group-normalized advantage, $\epsilon$ is the clipping range, and $\beta$ is the KL penalty coefficient.


\section{Experiments}

\subsection{Experimental Setup}

\paragraph{Implementation Details.}
We use Qwen3-VL-4B-Instruct and Qwen3-VL-8B-Instruct~\cite{bai2025qwen3vl} as the backbone models.
The reasoning and embedding LoRAs are applied to all linear projections in the language model with rank $r{=}32$, $\alpha{=}64$ for the 4B model and $r{=}64$, $\alpha{=}128$ for the 8B model.
We use $K{=}16$ learnable query tokens for embedding extraction and a routing gate $g_\phi$.
The total trainable parameters are 133M (3.3\% of backbone) for TWN-4B and 351M (4.6\%) for TWN-8B.
For the SFT stage, following VLM2Vec-V2~\cite{meng2025vlm2vec}, we use a loss temperature of $\tau{=}0.02$ and use GradCache~\cite{gao2021scaling} to scale the global batch size to $2{,}048$.
All loss weights in $\mathcal{L}_{\text{SFT}}$ are set to $\lambda_{\text{base}} = \lambda_{\text{cot}} = \lambda_{\text{route}} = 1$.
We train for 3 epochs with a learning rate of $5{\times}10^{-4}$.
For the reinforcement learning stage, we use GRPO~\cite{shao2024deepseekmath} with group size $G{=}8$, KL coefficient $\beta{=}0.1$, and a learning rate of $5{\times}10^{-6}$.
For reward computation, we randomly sample $2{,}048$ negatives from the global embedding cache $\mathcal{B}$.
Additional implementation details are provided in Appendix~\ref{sec:impl_details}.

\paragraph{Benchmarks and Baselines.}
We evaluate on MMEB-V2~\cite{meng2025vlm2vec}, a comprehensive benchmark covering 78 tasks across image, video, and visual document modalities.
Following prior work, we report Hit@1 for image and video tasks and NDCG@5~\cite{jarvelin2002cumulated} for visual document tasks.
We compare with two categories of methods.
\textit{Discriminative embedding} methods encode multimodal inputs into embeddings without reasoning:
ColPali~\cite{faysse2024colpali},
VLM2Vec~\cite{jiang2024vlm2vec},
GME~\cite{zhang2025bridging},
LamRA~\cite{liu2025lamra},
CAFe~\cite{yu2025cafe},
and VLM2Vec-V2~\cite{meng2025vlm2vec}.
\textit{Generative embedding} methods incorporate chain-of-thought reasoning to improve representation quality:
TTE~\cite{cui2025think}
and UME-R1~\cite{lan2025ume}.
We exclude models such as Qwen3-VL-Embedding and Seed-1.6-Embedding from our comparison, as they are trained on significantly larger proprietary corpora, making a direct comparison unfair.

\subsection{Main Results}

\begin{table}[htbp]
	\caption{Comparison of performance between baselines and our method on MMEB-V2. CLS: classification, QA: question answering, RET: retrieval, GD: grounding, MRET: moment retrieval, VDR: ViDoRe, VR: VisRAG, OOD: out-of-domain. The highest and second-highest values are highlighted in bold and underline.}
	\vspace{2mm}
	\centering
	\renewcommand{\arraystretch}{1.2}
	\resizebox{\textwidth}{!}{
		\begin{tabular}{l ccccc ccccc ccccc c}
			\toprule
			\multirow{2}{*}{Model}                   & \multicolumn{5}{c}{Image} & \multicolumn{5}{c}{Video} & \multicolumn{5}{c}{VisDoc} & \multirow{2}{*}{All}                                                                                                \\
			\cmidrule(lr){2-6} \cmidrule(lr){7-11} \cmidrule(lr){12-16}
			                                         & CLS                       & QA                        & RET                        & GD                   & Overall & CLS  & QA   & RET  & MRET & Overall & VDRv1 & VDRv2 & VR   & OOD  & Overall        \\
			\midrule
			\# of Datasets                           & 10                        & 10                        & 12                         & 4                    & 36      & 5    & 5    & 5    & 3    & 18      & 10    & 4     & 6    & 4    & 24      & 78   \\
			\midrule
			\rowcolor[HTML]{EDEDED}
			\multicolumn{17}{c}{\emph{Baseline Models}}                                                                                                                                                                                                         \\
			\midrule
			ColPali-V1.3-3B~\cite{faysse2024colpali} & 40.3                      & 11.5                      & 48.1                       & 40.3                 & 34.9    & 26.7 & 37.8 & 21.6 & 25.5 & 28.2    & 83.6  & 52.0  & 81.1 & 43.1 & 71.0    & 44.4 \\
			GME-2B~\cite{zhang2025bridging}          & 54.4                      & 29.9                      & 66.9                       & 55.5                 & 51.9    & 34.9 & 42.0 & 25.6 & 32.4 & 33.9    & \underline{86.1}  & 54.0  & 82.5 & 43.1 & 72.7    & 54.1 \\
			GME-7B~\cite{zhang2025bridging}          & 57.7                      & 34.7                      & 71.2                       & 59.3                 & 56.0    & 37.4 & 50.4 & 28.4 & 37.0 & 38.4    & \textbf{89.4}  & 55.6  & 85.0 & 44.4 & 75.2    & 57.8 \\
			LamRA-2-7B~\cite{liu2025lamra}           & 59.2                      & 26.5                      & 70.0                       & 62.7                 & 54.1    & 39.3 & 42.6 & 24.3 & 34.6 & 35.2    & 22.0  & 11.5  & 37.4 & 21.0 & 23.9    & 40.4 \\
			LamRA-2.5-7B~\cite{liu2025lamra}         & 51.7                      & 34.1                      & 66.9                       & 56.7                 & 52.4    & 32.9 & 42.6 & 23.2 & 37.6 & 33.7    & 56.3  & 33.3  & 58.2 & 40.1 & 50.2    & 47.4 \\
			VLM2Vec-2B~\cite{jiang2024vlm2vec}       & 58.7                      & 49.3                      & 65.0                       & 72.9                 & 59.7    & 33.4 & 30.5 & 20.6 & 33.0 & 29.0    & 49.8  & 13.5  & 51.8 & 33.5 & 41.6    & 47.0 \\
			VLM2Vec-7B~\cite{jiang2024vlm2vec}       & 62.7                      & 56.9                      & 69.4                       & 82.2                 & 65.5    & 39.1 & 30.0 & 29.0 & 38.9 & 33.7    & 56.9  & 9.4   & 59.1 & 38.1 & 46.4    & 52.3 \\
			VLM2Vec-V2~\cite{meng2025vlm2vec}        & 62.9                      & 56.3                      & 69.5                       & 77.3                 & 64.9    & 39.3 & 34.3 & 28.8 & 36.8 & 34.6    & 75.5  & 45.1  & 79.6 & 39.6 & 65.4    & 58.0 \\
			CAFe-7B~\cite{yu2025cafe}                & 63.6                      & 61.7                      & 69.1                       & \underline{87.6}                 & 67.6    & 35.8 & 58.7 & 34.4 & 39.5 & 42.4    & 70.7  & 49.6  & 79.5 & 38.1 & 63.9    & 60.6 \\
			UME-R1-2B~\cite{lan2025ume}              & 64.8                      & 62.8                      & 67.6                       & 77.2                 & 66.6    & 44.3 & 51.0 & 32.9 & \underline{39.7} & 42.2    & 72.4  & 46.2  & 79.2 & 37.2 & 63.9    & 60.1 \\
			UME-R1-7B~\cite{lan2025ume}              & 67.1                      & 69.2                      & \underline{71.9}                       & 84.9                 & 71.3    & 48.6 & 60.7 & \textbf{38.2} & 39.3 & \underline{47.5}    & 75.7  & 50.5  & 83.7 & 37.6 & 67.1    & 64.5 \\
			TTE$_s$-2B~\cite{cui2025think}           & 67.9                      & 66.6                      & 70.2                       & 84.1                 & 70.1    & 47.3 & 49.1 & 33.2 & 32.1 & 41.3    & 77.5  & 53.2  & 83.2 & 41.1 & 68.8    & 63.1 \\
			TTE$_s$-7B~\cite{cui2025think}           & \underline{69.7}                      & \underline{72.4}                      & \textbf{74.0}                       & \textbf{90.6}                 & \textbf{74.2}    & \underline{49.1} & 60.6 & \underline{36.4} & 37.2 & 46.8    & 84.1  & \textbf{62.7}  & \textbf{91.9} & 47.6 & \underline{76.4}    & \underline{68.6} \\
			\midrule
			\rowcolor[HTML]{EDEDED}
			\multicolumn{17}{c}{\emph{Ours}}                                                                                                                                                                                                                    \\
			\midrule
			TWN-4B (base)                            & 66.2                      & 67.6                      & 67.6                       & 85.9                 & 69.2    & 42.3 & 57.7 & 33.0 & 34.5 & 42.7    & 80.8  & 52.5  & 84.4 & 66.1 & 74.6    & 64.8 \\
			TWN-4B (cot)                             & 66.2                      & 69.2                      & 67.4                       & 85.5                 & 69.6    & 46.0 & 58.9 & 33.4 & 34.9 & 44.3    & 81.4  & 53.2  & 84.8 & 67.2 & 75.2    & 65.5 \\
			TWN-4B (adaptive)                        & 68.6                      & 71.7                      & 68.0                       & 87.0                 & 71.3    & 45.8 & \underline{63.4} & 33.6 & 36.6 & 45.7    & 81.5  & 53.6  & 84.8 & 67.3 & 75.3    & 66.6 \\
			\color{gray} TWN-4B (oracle$^\dagger$)   & \color{gray} 71.1         & \color{gray} 74.8         & \color{gray} 71.2          & \color{gray} 88.5    & \color{gray} 74.1 & \color{gray} 49.2 & \color{gray} 67.4 & \color{gray} 37.2 & \color{gray} 40.7 & \color{gray} 49.5 & \color{gray} 83.6 & \color{gray} 58.3 & \color{gray} 87.1 & \color{gray} 69.5 & \color{gray} 77.9 & \color{gray} 69.6 \\
			\cmidrule(lr){1-17}
			TWN-8B (base)                            & 68.6                      & 70.5                      & 69.8                       & 85.9                 & 71.5    & 46.7 & 58.0 & 33.0 & 35.5 & 44.1    & 80.9  & 52.6  & 85.2 & 68.8 & 75.2    & 66.3 \\
			TWN-8B (cot)                             & 68.3                      & 72.3                      & 69.4                       & 86.2                 & 71.8    & 48.8 & 61.8 & 33.8 & 39.4 & 46.7    & 81.9  & 55.8  & 85.7 & \underline{69.3} & 76.4    & 67.4 \\
			TWN-8B (adaptive)                        & \textbf{70.2}                      & \textbf{74.3}                      & 70.8                       & 87.1                 & \underline{73.4}    & \textbf{50.1} & \textbf{64.0} & 34.8 & \textbf{40.8} & \textbf{48.2}    & 82.5  & \underline{56.7}  & \underline{86.2} & \textbf{70.0} & \textbf{77.0}    & \textbf{68.7} \\
			\color{gray} TWN-8B (oracle$^\dagger$)   & \color{gray} 72.9         & \color{gray} 77.4         & \color{gray} 72.8          & \color{gray} 89.2    & \color{gray} 75.9 & \color{gray} 53.6 & \color{gray} 69.9 & \color{gray} 37.4 & \color{gray} 44.4 & \color{gray} 52.1 & \color{gray} 84.0 & \color{gray} 59.4 & \color{gray} 87.5 & \color{gray} 71.9 & \color{gray} 78.7 & \color{gray} 71.3 \\
			\bottomrule
			\multicolumn{17}{l}{\footnotesize $^\dagger$ Oracle selects the better of base/cot per sample (theoretical upper bound).}
		\end{tabular}
	}
	\label{tab:main_result}
\end{table}

We evaluate TWN under three inference modes: \emph{base} (discriminative only, $w{=}0$), \emph{cot} (always generate CoT, $w{=}1$), and \emph{adaptive} (the routing gate $g_\phi$ decides per input).
As shown in Table~\ref{tab:main_result}, TWN-8B with adaptive routing achieves the highest overall score of 68.7, surpassing recent state-of-the-art methods, and TWN-4B (66.6) also obtains competitive performance (detailed per-task scores are provided in Appendix~\ref{sec:detailed_scores_v2}).
For both scales, the adaptive mode consistently surpasses base and cot modes (4B: 66.6 vs.\ 64.8/65.5; 8B: 68.7 vs.\ 66.3/67.4), suggesting that base and CoT-enhanced embeddings are complementary, and the routing gate can dynamically select the more effective strategy per input (see Appendix~\ref{sec:case_studies} for qualitative examples).
The oracle upper bound, which picks the better of base/cot per sample, reaches 69.6 for TWN-4B and 71.3 for TWN-8B, indicating that improved routing can unlock further gains.

To compare inference efficiency, Table~\ref{tab:efficiency_tradeoff} reports the average number of reasoning tokens per input.
TWN-8B (adaptive) produces 62.8 tokens on average, a 45.4\% reduction compared to its cot mode (115.2 tokens), while achieving higher retrieval quality.
Compared to UME-R1-7B, which generates 458.7 tokens per input, TWN-8B (adaptive) uses $7.3\times$ fewer tokens. Two factors contribute to this reduction: our CoT construction pipeline explicitly discourages redundant reasoning, yielding compact traces, and the adaptive think mechanism skips CoT entirely for simple inputs. As a result, TWN achieves superior retrieval quality with significantly lower inference cost.

\begin{table}[!ht]
	\caption{Average number of reasoning tokens per input under each inference strategy, broken down by modality. The base (discriminative) mode generates zero reasoning tokens and is omitted. Reduction denotes the token saving of adaptive routing relative to the full CoT mode of the same model. The lowest and second-lowest values are highlighted in \textbf{bold} and \underline{underline}.}
	\vspace{1mm}
	\centering\small
	\begin{tabular}{l cccc c}
		\toprule
		\multirow{2}{*}{Method} & \multicolumn{4}{c}{\# Reasoning Tokens ($\downarrow$)} & \multirow{2}{*}{Reduction} \\
		\cmidrule(lr){2-5}
		                    & Image & Video & VisDoc & Avg   &           \\
		\midrule
		\rowcolor[HTML]{EDEDED}
		\multicolumn{6}{c}{\emph{Baselines}} \\
		\midrule
		UME-R1-2B           & 331.9 & 380.1 & 478.8  & 388.2 & --        \\
		UME-R1-7B           & 452.7 & 372.4 & 532.4  & 458.7 & --        \\
		\midrule
		\rowcolor[HTML]{EDEDED}
		\multicolumn{6}{c}{\emph{Ours}} \\
		\midrule
		TWN-4B (cot)        & 74.4  & 114.7 & 183.1  & 117.2 & --        \\
		TWN-4B (adaptive)   & \textbf{42.3}  & \textbf{62.1}  & \textbf{81.3}   & \textbf{58.9}  & $-$49.7\% \\
		\cmidrule(lr){1-6}
		TWN-8B (cot)        & 72.5  & 125.1 & 171.7  & 115.2 & --        \\
		TWN-8B (adaptive)   & \underline{44.6}  & \underline{71.8}  & \underline{83.5}   & \underline{62.8}  & $-$45.4\% \\
		\bottomrule
	\end{tabular}
	\label{tab:efficiency_tradeoff}
\end{table}

\subsection{Ablation Study}

We ablate the architecture, RL training, and adaptive routing of TWN using the 4B backbone. All variants share the same training data and hyperparameters unless otherwise noted.

\begin{table}[!t]
	\caption{Ablation study on TWN-4B. (a)~Architecture variants are compared under the same SFT configuration. The best SFT variant (Dual-LoRA w/ detach) serves as the starting point for (b)~RL ablations. All variants use the adaptive routing strategy for evaluation.}
	\vspace{1mm}
	\centering\small
	\begin{tabular}{lcccc}
		\toprule
		Configuration        & Image & Video & VisDoc & All     \\
		\midrule
		\rowcolor[HTML]{EDEDED}
		\multicolumn{5}{l}{\emph{(a) Architecture (SFT stage)}} \\
		\midrule
		Shared LoRA          & 68.0  & 41.5  & 72.6   & 63.3    \\
		Dual-LoRA w/o detach & 68.9  & 42.5  & 73.2   & 64.1    \\
		Dual-LoRA w/ detach  & \textbf{69.8}  & \textbf{44.0}  & \textbf{74.2}   & \textbf{65.2}    \\
		\midrule
		\rowcolor[HTML]{EDEDED}
		\multicolumn{5}{l}{\emph{(b) Reinforcement Learning}}   \\
		\midrule
		w/o RL (= SFT only)  & 69.8  & 44.0  & 74.2   & 65.2    \\
		+ RL (in-batch neg.) & 70.2  & 44.4  & 74.5   & 65.6    \\
		+ RL (global cache)  & \textbf{71.3}  & \textbf{45.7}  & \textbf{75.3}   & \textbf{66.6}    \\
		\bottomrule
	\end{tabular}
	\label{tab:ablation}
\end{table}

\paragraph{Architecture (SFT Stage).}
Table~\ref{tab:ablation}(a) compares three architecture variants under the same SFT configuration, progressively increasing the degree of gradient separation.
A Shared LoRA that handles both reasoning and embedding reaches 63.3.
Separating into two adapters (Dual-LoRA w/o detach) improves to 64.1, and further detaching gradients at the adapter boundary (Dual-LoRA w/ detach) yields the best result of 65.2.
The consistent improvement as the degree of gradient separation increases suggests that gradient conflict between the generative and discriminative objectives is a key factor affecting performance (training dynamics are visualized in Appendix~\ref{sec:training_dynamics}).
Looking at per-modality gains from the detachment step, Video benefits the most ({+}1.5), followed by VisDoc ({+}1.0) and Image ({+}0.9).

\paragraph{Reinforcement Learning.}
Table~\ref{tab:ablation}(b) ablates the RL stage, starting from the best SFT variant (65.2).
RL with in-batch negatives brings a modest gain of {+}0.4, while switching to the global cache $\mathcal{B}$ yields a substantially larger improvement of {+}1.4 (66.6).
The contrast ({+}0.4 vs.\ {+}1.4) suggests that a larger and more diverse negative pool provides higher-quality reward signals that further improve CoT generation.
Across modalities, the global cache brings consistent gains (Image {+}1.5, Video {+}1.7, VisDoc {+}1.1).

\begin{figure}[!ht]
	\centering
	\includegraphics[width=\textwidth]{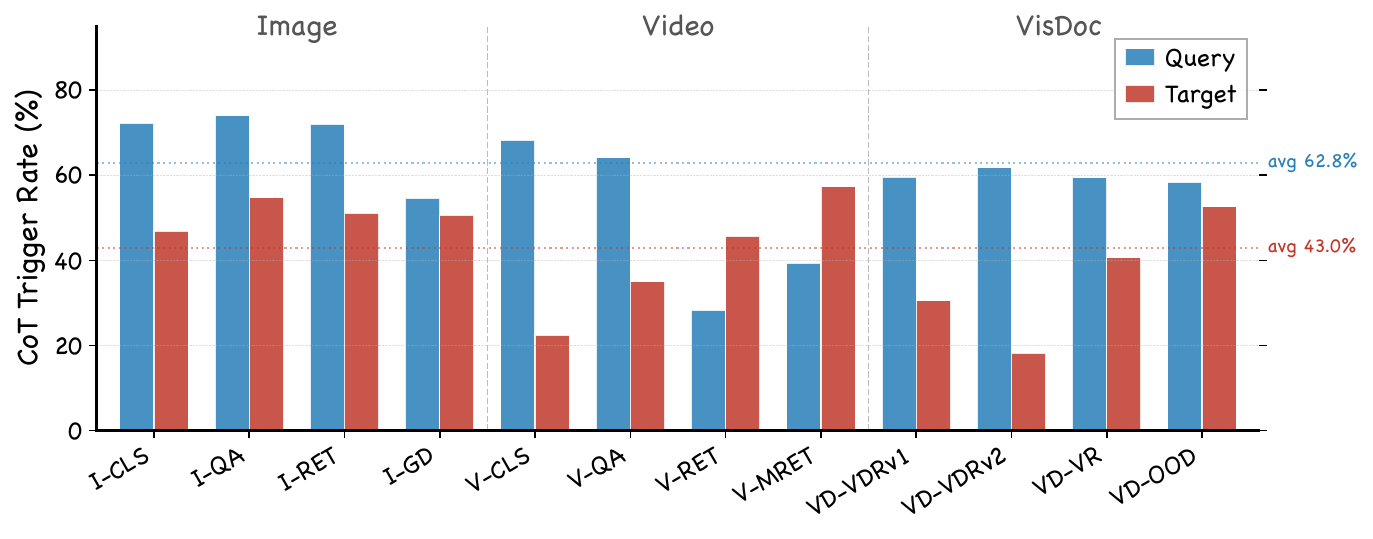}
	\caption{Per-category CoT trigger rate (\%) of the adaptive routing strategy. Blue bars show query-side rates; red bars show target-side rates. Dashed lines indicate overall averages.}
	\label{fig:cot_trigger_rate}
\end{figure}

\paragraph{Routing Strategy.}
Figure~\ref{fig:cot_trigger_rate} reports the per-category CoT trigger rate, i.e., the percentage of inputs routed to CoT generation (higher means more reasoning).
Overall, the query side exhibits a higher trigger rate (62.8\%) than the target side (43.0\%), reflecting the asymmetric complexity between queries and targets: queries often involve complex intent that benefits from reasoning, while targets (e.g., images, document pages) tend to be more self-contained.
Across task categories, tasks requiring deeper semantic understanding (e.g., QA) tend to have higher trigger rates, while more straightforward tasks (e.g., grounding) show lower rates.
This indicates that the routing gate learns to allocate reasoning resources in a task-aware manner without explicit supervision.

\section{Related Work}

\paragraph{Multimodal Embedding Models.}
Early vision-language models such as CLIP~\cite{radford2021learning}, ALIGN~\cite{jia2021align}, and SigLIP~\cite{zhai2023sigmoid} established the dual-encoder paradigm through large-scale image-text contrastive pre-training but are limited to simple paired inputs.
Recent efforts repurpose MLLMs~\cite{liu2023llava,li2023blip2,chen2024internvl,wang2024qwen2vl} as embedding backbones: E5-V~\cite{jiang2024e5v}, VLM2Vec~\cite{jiang2024vlm2vec}, and MM-Embed~\cite{lin2024mmembed} fine-tune MLLMs with contrastive objectives, substantially outperforming dual-encoder methods.
Subsequent work addresses data and training challenges through automated corpus construction~\cite{zhou2024megapairs,zhang2025bridging}, improved negative sampling~\cite{lan2025llave,gu2025unime}, and hybrid contrastive-autoregressive losses~\cite{yu2025cafe,ouali2024vladva}.
Concurrently, text embedding research has developed techniques---instruction-aware training~\cite{su2023instructor,wang2024improving}, weakly-supervised contrastive pre-training~\cite{wang2022text}, and bidirectional adaptation of autoregressive LLMs~\cite{behnamghader2024llm2vec,lee2024nvembed}---that inform how MLLMs are repurposed as encoders.
VLM2Vec~\cite{jiang2024vlm2vec} introduces the MMEB benchmark---a multimodal counterpart of MTEB~\cite{muennighoff2023mteb}---spanning visual classification, question answering, retrieval, and grounding. Its successor, MMEB-V2~\cite{meng2025vlm2vec}, further covers video and visual-document modalities.
Despite this rapid progress, these approaches treat the MLLM purely as a discriminative encoder, neglecting its generative and reasoning capabilities.

\paragraph{Reasoning-Enhanced Multimodal Embeddings.}
Recent work incorporates reasoning into the embedding pipeline to improve retrieval performance for semantically complex inputs~\cite{liu2025lamra}.
These methods fall into two broad categories. \emph{Decoupled} designs, exemplified by TTE~\cite{cui2025think}, employ a dedicated MLLM reasoner to produce CoT~\cite{wei2022chain,kojima2022large} traces that a separate embedder consumes alongside the original input. This separation sidesteps gradient conflicts but nearly doubles the parameter budget. \emph{Joint} designs, such as RGE~\cite{liu2025rge} and UME-R1~\cite{lan2025ume}, let a single model handle both reasoning and embedding, achieving parameter efficiency but potentially introducing gradient conflicts between the autoregressive and contrastive objectives.
TWN proposes a shared-backbone dual-LoRA architecture that combines the strengths of both camps: two lightweight adapters sit on the same frozen backbone, keeping parameters close to a single model while isolating the two gradient flows. Beyond this architectural choice, existing methods also generate CoT indiscriminately for every input; TWN addresses this with an adaptive think mechanism that selectively invokes reasoning on a per-input basis.

\paragraph{Reinforcement Learning for Embedding.}
RL has pushed LLM reasoning beyond supervised fine-tuning, as demonstrated by DeepSeek-R1~\cite{deepseek2025r1}. GRPO~\cite{shao2024deepseekmath} eliminates the need for a critic network through group-relative advantage estimation.
UME-R1~\cite{lan2025ume} first applied GRPO to embeddings, using embedding quality as the reward, but uses a single shared model for both reasoning and reward computation.
TWN freezes the embedding LoRA as a stable reward environment and introduces a global embedding cache for more discriminative reward signals.


\section{Conclusion}

We presented Think When Needed (TWN), a unified framework for reasoning-driven multimodal embeddings that addresses two fundamental limitations of existing generative embedding methods: gradient conflicts between reasoning and embedding objectives, and the indiscriminate application of chain-of-thought reasoning regardless of input complexity.
TWN introduces three key components: (1)~a dual-LoRA architecture that attaches separate reasoning and embedding adapters to a shared frozen backbone with gradient detachment, mitigating cross-objective interference while maintaining parameter efficiency; (2)~an adaptive think mechanism with a self-supervised routing gate that adaptively selects between discriminative and generative embeddings per input; and (3)~embedding-guided RL that exploits dual-LoRA's parameter separation to freeze the embedding adapter as a stable reward environment, coupled with a global embedding cache for more discriminative reward signals.
On the 78 tasks of MMEB-V2, TWN achieves state-of-the-art retrieval performance, while reducing reasoning tokens by up to 50\% compared to the full generative mode.
These results suggest that adaptive reasoning allocation can simultaneously improve both the quality and efficiency of multimodal embeddings.

\newpage
\bibliographystyle{plainnat}

\newpage
\appendix


\section{Implementation Details}
\label{sec:impl_details}

\paragraph{Multimodal Input Processing.}
Images are processed at resolutions from $4{,}096$ to $1{,}048{,}576$ pixels, corresponding to $4$ to $1{,}024$ visual tokens with a patch size of $32{\times}32$.
Videos are uniformly sampled to 8 frames, with at most $524{,}288$ pixels per frame, corresponding to $512$ visual tokens with a patch size of $32{\times}32$.
The maximum sequence length is set to $8{,}192$ tokens for all stages.

\paragraph{Training Setup.}
All experiments are conducted on 32 NVIDIA H20 GPUs with DeepSpeed ZeRO-2~\cite{rasley2020deepspeed}, FlashAttention-2~\cite{dao2023flashattention2}, and BF16 mixed precision.
Both the reasoning and embedding LoRA adapters are applied to all linear projection layers (\texttt{q\_proj}, \texttt{k\_proj}, \texttt{v\_proj}, \texttt{o\_proj}, \texttt{gate\_proj}, \texttt{up\_proj}, \texttt{down\_proj}) in the language model, excluding all visual encoder modules. For the 4B model, both adapters use rank $r{=}32$ and scaling factor $\alpha{=}64$; for the 8B model, rank $r{=}64$ and $\alpha{=}128$. All adapters use zero dropout.
We use $K{=}16$ learnable query tokens for embedding extraction. The routing gate $g_\phi$ takes the last input token's hidden state as input.
All stages are optimized with AdamW ($\beta_1{=}0.9$, $\beta_2{=}0.999$, weight decay $1{\times}10^{-4}$).

In Stage~1 (SFT), we use a learning rate of $5{\times}10^{-4}$ with $5\%$ linear warmup and train for 3 epochs.

In Stage~2 (RL), we initialize the RL policy adapter from the Stage-1 reasoning adapter weights and freeze all other components (embedding adapter, learnable queries, routing gate), providing a stationary reward environment.
All target embeddings are pre-computed into a static global cache from the SFT checkpoint before RL training begins.
We use GRPO with group size $G{=}8$, KL coefficient $\beta{=}0.1$, a sampling temperature of $1.0$, a learning rate of $5{\times}10^{-6}$, and gradient clipping at max norm $1.0$.
The maximum CoT generation length is $2{,}048$ tokens.
The gap reward samples $2{,}048$ negatives from the global cache with a softmax temperature of $\tau_r{=}0.1$ for hard-negative weighting.
We train for 1 epoch.

\paragraph{Parameter Breakdown.}
Table~\ref{tab:param_breakdown} summarizes the parameter counts of TWN\@.
The frozen backbone accounts for the vast majority of parameters (4.02B for the 4B model and 7.57B for the 8B model), while the total trainable parameters amount to only 133M (3.3\% of backbone) and 351M (4.6\%), respectively.
The routing gate and learnable queries together contribute fewer than 1.4M/2.2M parameters ($<$0.04\% of backbone), adding negligible overhead in terms of model size.
The marginal cost of adding the second adapter over a single-LoRA baseline is 67M/177M, corresponding to only 1.7\%/2.3\% of the backbone.

\begin{table}[!htbp]
	\caption{Parameter breakdown of TWN\@.}
	\vspace{1mm}
	\centering\small
	\begin{tabular}{lcc}
		\toprule
		Component                       & Qwen3-VL-4B       & Qwen3-VL-8B       \\
		\midrule
		Backbone (frozen)               & 4.02B             & 7.57B             \\
		Dual LoRA ($\times 2$ adapters) & 132.1M ($r{=}32$) & 349.2M ($r{=}64$) \\
		Learnable queries ($K{=}16$)    & 41.0K             & 65.5K             \\
		Routing gate                    & 1.3M              & 2.1M              \\
		\midrule
		Total trainable                 & 133.5M (3.3\%)    & 351.4M (4.6\%)    \\
		\bottomrule
	\end{tabular}
	\label{tab:param_breakdown}
\end{table}

\section{Dataset Construction}
\label{sec:dataset_construction}

\subsection{Data Sources}

We construct a comprehensive training dataset from three primary sources spanning image, video, and visual document modalities, following the data paradigm of VLM2Vec-V2~\cite{meng2025vlm2vec}.

\paragraph{Image-based tasks.}
We adopt 20 datasets from the MMEB training splits, covering four meta-task categories: classification (ImageNet-1K, N24News, HatefulMemes, VOC2007, SUN397), visual question answering (OK-VQA, A-OKVQA, DocVQA, InfographicsVQA, ChartQA, Visual7W), retrieval (VisDial, CIRR, VisualNews, MSCOCO, NIGHTS, WebQA), and visual grounding (RefCOCO), where VisualNews and MSCOCO each include both text-to-image and image-to-text retrieval directions, yielding 20 distinct training tasks in total.

\paragraph{Video-based tasks.}
We incorporate LLaVA-Hound video instruction data, including video captioning (used bidirectionally for text-to-video and video-to-text retrieval) and video question answering, to enable video understanding capabilities.

\paragraph{Visual document tasks.}
We include ViDoRe, VisRAG in-domain, and VisRAG synthetic training data for visual document retrieval.

The training set comprises a total of $\sim$1.77M query-target pairs across all three modalities.
Table~\ref{tab:dataset_stats} summarizes the per-dataset statistics, including the number of training samples, both-side CoT quality, and the modality pattern of each dataset.

\begin{table}[!htbp]
	\caption{Statistics of training data with CoT annotations.
		\textbf{Clean}: percentage of samples where both query-side and target-side CoT judgments pass (correct or exempt).
		\textbf{Modality}: input modality pattern (T: text, I: image, V: video, D: document).}
	\vspace{2mm}
	\centering
	\small
	\setlength{\tabcolsep}{12pt}
	\renewcommand{\arraystretch}{1.05}
	\begin{tabular}{l r r c}
		\toprule
		\textbf{Dataset} & \textbf{\#Pairs} & \textbf{Clean (\%)} & \textbf{Modality} \\
		\midrule
		\rowcolor[HTML]{EDEDED}
		\multicolumn{4}{c}{\emph{Image-based (MMEB-train, 20 datasets)}} \\
		ImageNet-1K       &  99,984 &  58.0 & T+I $\to$ T   \\
		N24News           &  48,979 &  37.6 & T+I $\to$ T   \\
		HatefulMemes      &   8,500 &  73.1 & T+I $\to$ T   \\
		VOC2007           &   7,844 &  68.8 & T+I $\to$ T   \\
		SUN397            &  19,835 &  69.0 & T+I $\to$ T   \\
		OK-VQA            &   8,988 &  61.0 & T+I $\to$ T   \\
		A-OKVQA           &  17,016 &  61.8 & T+I $\to$ T   \\
		DocVQA            &  39,198 &  97.2 & T+I $\to$ T   \\
		InfographicsVQA   &  23,732 &  90.7 & T+I $\to$ T   \\
		ChartQA           &  28,155 &  81.2 & T+I $\to$ T   \\
		Visual7W          &  69,765 &  71.8 & T+I $\to$ T   \\
		VisDial           & 123,113 &  98.8 & T $\to$ T+I   \\
		CIRR              &  26,107 &  99.3 & T+I $\to$ T+I \\
		VisualNews-t2i    &  99,847 &  99.6 & T $\to$ T+I   \\
		VisualNews-i2t    &  99,913 &  99.2 & T+I $\to$ T   \\
		MSCOCO-t2i        &  99,976 &  99.9 & T $\to$ T+I   \\
		MSCOCO-i2t        & 112,831 &  99.9 & T+I $\to$ T   \\
		NIGHTS            &  15,940 &  99.8 & T+I $\to$ T+I \\
		WebQA             &  17,146 &  99.1 & T $\to$ T+I   \\
		MSCOCO            &  99,348 &  96.6 & T+I $\to$ T+I \\
		\midrule
		\rowcolor[HTML]{EDEDED}
		\multicolumn{4}{c}{\emph{Video-based (LLaVA-Hound)}} \\
		LLaVA-Hound-t2v   &  90,132 &  99.9 & T $\to$ V     \\
		LLaVA-Hound-v2t   &  90,132 &  99.9 & V $\to$ T     \\
		LLaVA-Hound-VQA   & 196,124 &  80.1 & V+T $\to$ T   \\
		\midrule
		\rowcolor[HTML]{EDEDED}
		\multicolumn{4}{c}{\emph{Document-based (ViDoRe and VisRAG)}} \\
		ViDoRe            & 116,626 &  88.8 & T $\to$ D     \\
		VisRAG-InDomain   & 122,414 &  98.9 & T $\to$ D     \\
		VisRAG-Synthetic  &  83,464 &  99.5 & T $\to$ D     \\
		\midrule
		\textbf{Image}    & 1,066,217 & 88.0 & --- \\
		\textbf{Video}    &   376,388 & 89.6 & --- \\
		\textbf{Document} &   322,504 & 95.4 & --- \\
		\midrule
		\textbf{Total}    & 1,765,109 & 89.7 & --- \\
		\bottomrule
	\end{tabular}
	\label{tab:dataset_stats}
\end{table}

\subsection{CoT Generation and Filtering}

\paragraph{Generation.}
We generate chain-of-thought annotations for both query and target sides of each training pair using Qwen3.5-35B-A3B as the annotation model.
All prompts share a common two-step structure: (1)~\textbf{Analyze (Think)}: produce step-by-step reasoning---brief or empty for simple inputs, detailed for complex ones; (2)~\textbf{Synthesize (Answer)}: produce a concise summary for retrieval.
The output is a structured JSON with \texttt{think} and \texttt{answer} fields, formatted as \texttt{<think>...</think><answer>...</answer>} and appended to each data sample.
Crucially, the Synthesize instruction is \textbf{task-specific}: we customize it per task category to guide the annotation model toward summaries suited to each downstream retrieval objective (Table~\ref{tab:task_prompts}).
For example, classification tasks instruct the model to output the most specific category label, VQA tasks require direct concise answers, and retrieval tasks ask for key semantic elements useful for cross-modal matching.
Target-side prompts are similarly specialized into four modality-specific templates: label interpretation (classification), visual content description (images), document layout analysis (documents), and temporal event description (videos); datasets without a registered target prompt use the default template.
Across the full training set, 85.1\% of queries and 45.8\% of targets receive non-empty reasoning traces, reflecting the asymmetric complexity between query and target sides.
The base generation template is shown below.

\begin{figure}[H]
	\centering
	\includegraphics[width=\textwidth]{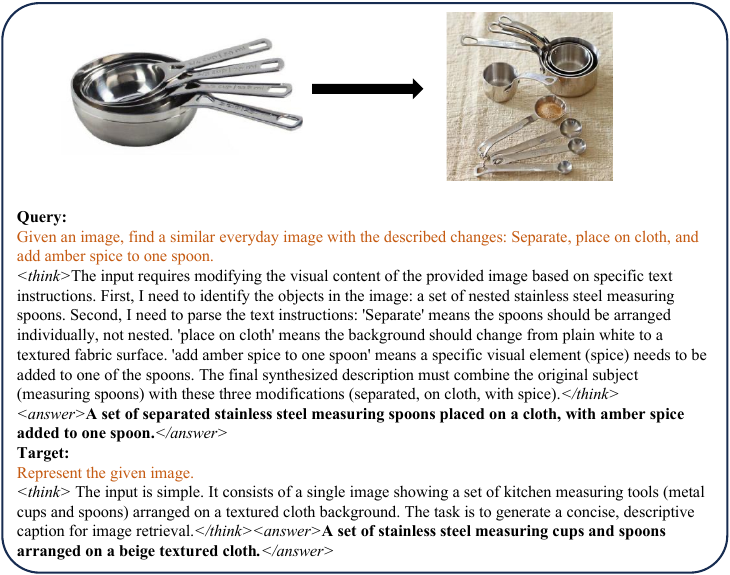}
	\caption{Training sample from the composed image retrieval task (CIRR). The query combines a reference image with a textual modification instruction, and the CoT trace decomposes the modification intent before producing a retrieval-oriented summary. The target side receives a concise visual description without reasoning, as the content is semantically straightforward.}
	\label{fig:training_example_1}
\end{figure}

\begin{figure}[H]
	\centering
	\includegraphics[width=\textwidth]{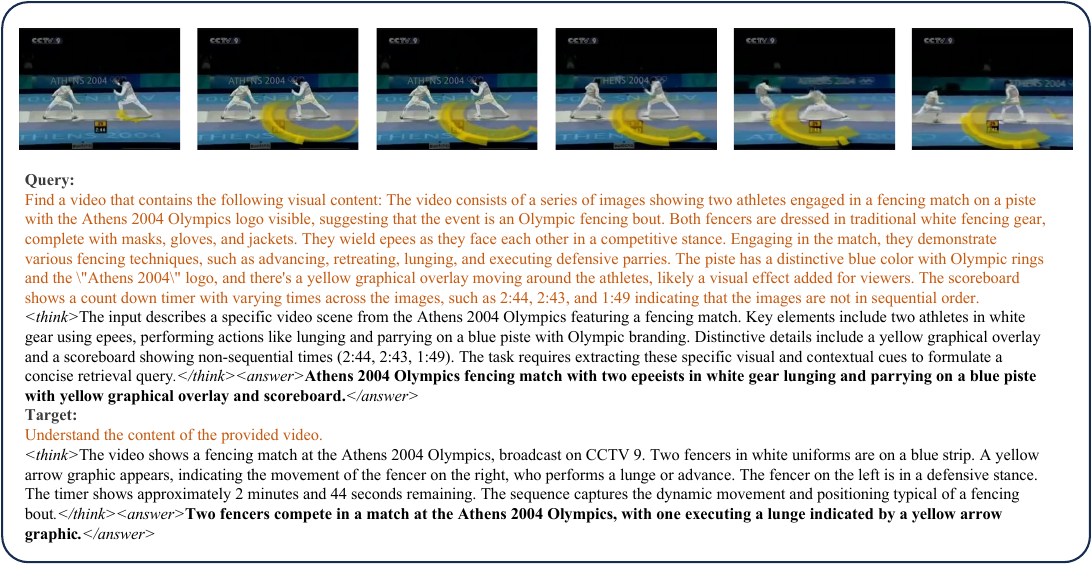}
	\caption{Training sample from the video retrieval task. The query provides a detailed textual description of a video scene, and the CoT trace extracts key visual and contextual cues for retrieval. The target side generates reasoning over the sampled video frames to produce a descriptive summary capturing the temporal dynamics and scene content.}
	\label{fig:training_example_2}
\end{figure}

\begin{figure}[H]
	\centering
	\includegraphics[width=\textwidth]{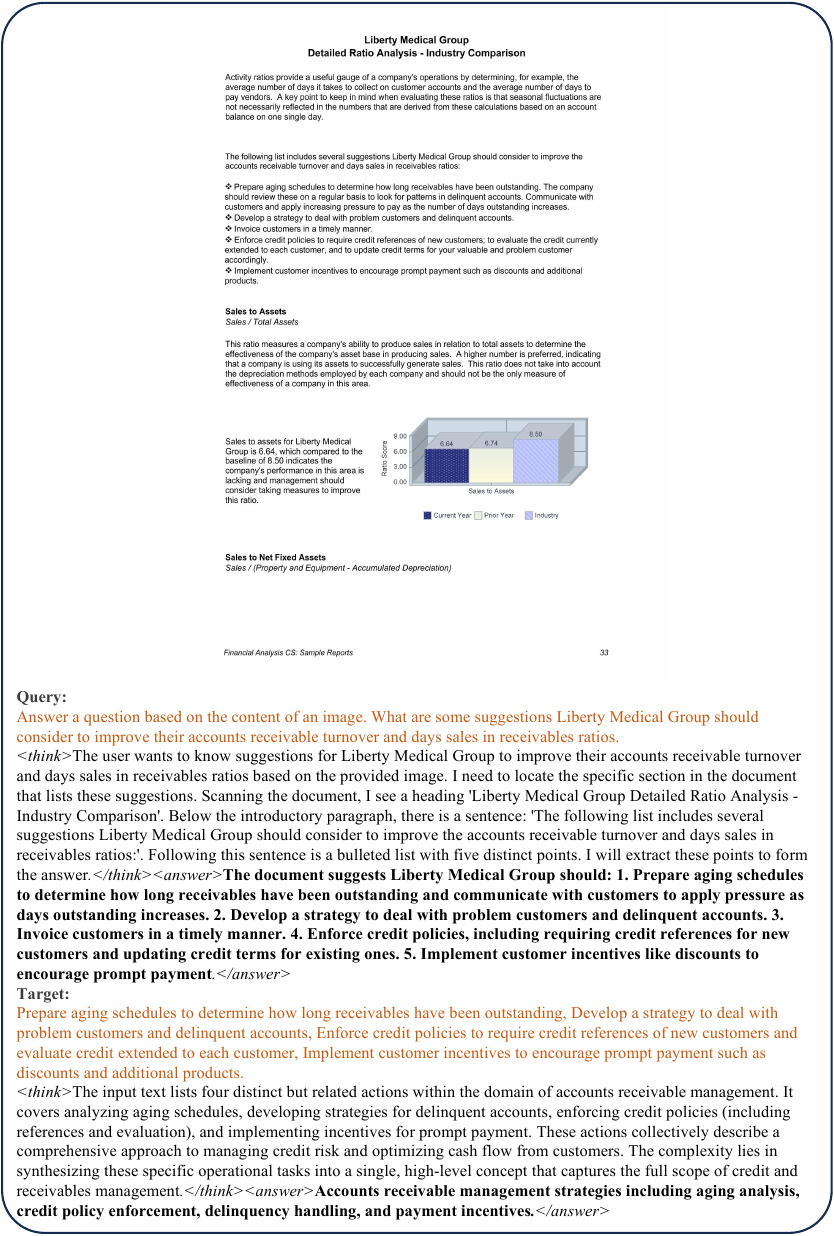}
	\caption{Training sample from the visual document question answering task (DocVQA). The query asks a question about a scanned financial report, and the CoT trace demonstrates document layout comprehension---locating relevant sections, parsing tabular data, and extracting structured information. The target-side CoT synthesizes the ground-truth answer text into a high-level semantic summary suitable for retrieval matching.}
	\label{fig:training_example_3}
\end{figure}

\paragraph{Filtering.}
To assess CoT quality, we employ Qwen3.5-35B-A3B as a judge model with \textbf{three task-adaptive validation modes}, assigned based on whether the task has a well-defined expected answer (Table~\ref{tab:task_prompts}).
(1)~\textbf{Strict verification} evaluates both reasoning quality (logical consistency, no hallucinations) and strict answer matching against the ground-truth target; it is applied to classification, VQA, document QA, and video QA queries.
(2)~\textbf{Hallucination-only verification} evaluates only whether the reasoning is free of hallucinations and the answer is relevant, without requiring answer matching; it is applied to retrieval and grounding tasks on both query and target sides, where no single correct summary exists.
(3)~\textbf{Skip}: no validation is performed for simple target-side inputs (e.g., classification labels, short VQA answers) where the CoT is typically trivial or empty.
The judge assigns a binary label (\texttt{is\_correct}: true/false) along with a brief justification; samples with empty CoT are treated as clean by default.
A training sample is considered clean when all non-skipped sides pass their respective validation.
After filtering, the overall clean rate is 89.7\%.
As shown in Table~\ref{tab:dataset_stats}, the clean rate varies significantly across datasets:
document-oriented and retrieval tasks (e.g., DocVQA at 97.2\%, VisRAG at 99.5\%) exhibit high CoT quality, while certain classification tasks with ambiguous or fine-grained labels (e.g., N24News at 37.6\%, ImageNet-1K at 58.0\%) show lower clean rates.
The judgment prompts for both verification modes are shown in Figures~\ref{fig:prompt_generation}, \ref{fig:prompt_strict}, and~\ref{fig:prompt_hallucination}.
Figures~\ref{fig:training_example_1}, \ref{fig:training_example_2}, and~\ref{fig:training_example_3} present representative training samples from our dataset, illustrating the structure of query-target pairs along with their generated CoT annotations across different task types and modalities.

\begin{table}[H]
	\caption{Task-specific answer instructions and validation configuration. All generation prompts share the same two-step Think--Answer structure (base template shown below); only the \textbf{Answer} instruction is customized per task category. Target-side prompts use four modality-specific templates: label interpretation (classification), visual content description (images), document layout analysis (documents), and temporal event description (videos). \textbf{Q/T}: validation mode for query/target side---\textbf{S}\,=\,strict, \textbf{H}\,=\,hallucination-only, --\,=\,skip.}
	\vspace{2mm}
	\centering\small
	\setlength{\tabcolsep}{5pt}
	\begin{tabular}{@{}p{0.25\textwidth}p{0.55\textwidth}cc@{}}
		\toprule
		\textbf{Task Category} & \textbf{Query-Side Answer Instruction} & \textbf{Q} & \textbf{T} \\
		\midrule
		Classification & Dataset-specific label: fine-grained category (ImageNet), object class (VOC2007), scene name (SUN397), news section (N24News), or Yes/No (HatefulMemes) & S & -- \\
		VQA & Direct, concise answer to the visual question & S & -- \\
		\addlinespace[3pt]
		Text$\to$Image Retrieval & Key visual elements, objects, scenes, or entities described in the text for image retrieval & H & H \\
		Image$\to$Text Retrieval & Concise description or caption of the image content for text matching & H & H \\
		Dialogue Retrieval & Combine key visual elements confirmed across multi-turn dialogue (VisDial) & H & H \\
		Composed Retrieval & Combine image content with text modification instructions to describe the target image (CIRR) & H & H \\
		Visual Similarity & Visual features, objects, colors, and composition for similar image retrieval (NIGHTS) & H & H \\
		Knowledge Retrieval & Key entities, concepts, and visual elements from the question for image retrieval (WebQA) & H & H \\
		\addlinespace[3pt]
		Visual Grounding & Target region attributes only; no background or scene context (RefCOCO) & H & H \\
		\addlinespace[3pt]
		Document QA & Direct answer from document content (ViDoRe) & S & -- \\
		Document Retrieval & Key visual and factual elements for document page retrieval (VisRAG) & H & H \\
		\addlinespace[3pt]
		Video QA & Direct answer based on video content & S & -- \\
		Text$\to$Video Retrieval & Key visual events, actions, and scenes from text for video retrieval & H & H \\
		\bottomrule
	\end{tabular}
	\label{tab:task_prompts}
\end{table}

\begin{figure}[!htbp]
	\begin{promptbox}{CoT Generation Prompt (Base Template)}
		\begin{lstlisting}[style=promptstyle]
You are an expert AI data annotator for a Multimodal Embedding Model.

**Input Data:**
{input_data}

**Task:**
1. **Analyze (Think):**
   - For **simple** inputs, provide a brief analysis or simply leave the "think" field completely empty ("").
   - For **complex** inputs, conduct a detailed, step-by-step analysis.
   - **Note:** Do not state whether the input is "simple" or "complex" in the "think" field; focus directly on the analysis itself.
2. **Synthesize (Answer):** Synthesize your analysis and reflection into a concise sentence or words that best captures the essence of the input for retrieval purposes.

**Output Format:**
Directly return a valid, raw JSON object.
{
  "think": "...",
  "answer": "..."
}
\end{lstlisting}
	\end{promptbox}
	\caption{Base template for CoT generation prompts. The Think step produces step-by-step reasoning, and the Answer step synthesizes a concise retrieval-oriented summary.}
	\label{fig:prompt_generation}
\end{figure}

\begin{figure}[!htbp]
	\begin{promptbox}{Strict Verification Prompt}
		\begin{lstlisting}[style=promptstyle]
You are an expert AI Judge. Your task is to evaluate whether the reasoning process is sound and if the answer matches the target.

**Task:**
Evaluate the **Reasoning to be Judged** and **Answer to be Judged** based on two strict criteria:
1. **Reasoning Quality:** The reasoning must be free of hallucinations and logically consistent with the **Original Query**.
2. **Answer Matching:** The **Answer** must be essentially the same as the **Target** --- only trivial surface-level differences are allowed.
If **Reasoning to be Judged** is empty or missing, judge ONLY based on the strict matching of the Answer.

**Input Data:**
- **Original Query:** {query}
- **Target:** {target}
- **Reasoning to be Judged:** {generated_think}
- **Answer to be Judged:** {generated_answer}

**Judgment Guidelines:**
1. **Correct (True):**
   * **Reasoning:** The reasoning is logical, directly addresses the query, and does not contain hallucinations or fabricated details not implied by the context.
   * **Answer:** The answer must refer to the exact same entity/concept as the **Target**. Only trivial differences are acceptable: exact synonyms (e.g., "cab" vs "taxi"), singular/plural forms, article/determiner differences, minor capitalization/punctuation.
2. **Incorrect (False):**
   * **Reasoning Flaws:** The reasoning contradicts the query, contains hallucinations, or is irrelevant.
   * **Answer Mismatch:** Overly generic hypernyms (e.g., "animal" for "dog"), related but different terms (e.g., "kitchen" for "oven"), or different entities/attributes/concepts.
   * **Ambiguity:** The answer is "I don't know", "Unclear", or the reasoning is nonsensical.

**Output Format:**
Directly return a valid, raw JSON object.
{
  "reason": "...",
  "is_correct": true/false
}
\end{lstlisting}
	\end{promptbox}
	\caption{Strict verification prompt for CoT quality judgment. Evaluates both reasoning quality and strict answer matching against the ground-truth target.}
	\label{fig:prompt_strict}
\end{figure}

\begin{figure}[!htbp]
	\begin{promptbox}{Hallucination-Only Verification Prompt}
		\begin{lstlisting}[style=promptstyle]
You are an expert AI Judge. Your task is to evaluate whether the reasoning process is sound and free of hallucinations.

**Task:**
Evaluate whether the **Reasoning to be Judged** and **Answer to be Judged** are free of hallucinations and logically consistent with the **Original Query**.
If **Reasoning to be Judged** is empty or missing, judge ONLY based on whether the **Answer** is reasonable and relevant.
If the Original Query contains retrieval or grounding intent (e.g., "Find", "Retrieve", "Represent"), the Answer is expected to be a descriptive text representation or keyword summary, not an actual retrieved item or factual answer. Keywords and phrases are acceptable as long as they are relevant for retrieval.

**Input Data:**
- **Original Query:** {query}
- **Reasoning to be Judged:** {generated_think}
- **Answer to be Judged:** {generated_answer}

**Judgment Guidelines:**
1. **Correct (True):**
   * **Reasoning:** The reasoning is logical, directly addresses the query, and does not contain hallucinations or fabricated details.
   * **Answer:** The answer is a reasonable and relevant description or response to the Original Query.
2. **Incorrect (False):**
   * **Reasoning Flaws:** The reasoning contradicts the query, contains hallucinations, or is irrelevant.
   * **Ambiguity:** The answer is nonsensical or completely unrelated.

**Output Format:**
Directly return a valid, raw JSON object.
{
  "reason": "...",
  "is_correct": true/false
}
\end{lstlisting}
	\end{promptbox}
	\caption{Hallucination-only verification prompt for CoT quality judgment. Evaluates only reasoning soundness and answer relevance, without requiring strict answer matching.}
	\label{fig:prompt_hallucination}
\end{figure}

\clearpage

\clearpage

\subsection{RL Data Sampling}

We construct the RL training set from the full SFT training data in two steps.

\paragraph{Step 1: Modality-Balanced Sampling.}
We sample ${\sim}20$K instances from the complete training set, equally divided across three modalities (image, video, and visual document, ${\sim}6{,}666$ each), with quotas distributed uniformly among datasets within each modality.
To prevent false negatives caused by duplicate target texts in contrastive training, classification datasets are deduplicated by unique target label, retaining at most one sample per class.
We exclude datasets whose targets are binary labels (HatefulMemes) or that have images on both the query and target sides (CIRR, NIGHTS, MSCOCO grounding), as these do not fit the RL paradigm where the query side benefits from reasoning while the target side is semantically straightforward.

\paragraph{Step 2: Embedding-Variance Filtering.}
Not all samples benefit equally from chain-of-thought reasoning.
To identify instances where reasoning quality meaningfully affects the resulting embedding, we generate $8$ independent CoT rollouts per sample using the SFT checkpoint and compute the pairwise cosine similarity among the $8$ corresponding embeddings.
Samples whose embeddings are highly consistent across rollouts---indicating that the embedding is insensitive to CoT content---are filtered out.
We rank samples by embedding variance within each modality and retain the top ${\sim}3{,}333$ per modality, yielding a final RL training set of ${\sim}10$K instances while preserving modality balance.

This filtering strategy ensures that RL optimization focuses on samples where improving CoT quality can substantively change the embedding representation, enabling GRPO to learn more efficiently.
Table~\ref{tab:rl_data} reports per-dataset statistics before and after filtering.

\begin{table}[H]
	\caption{RL training data statistics. \textbf{Pre-sample}: number of samples after modality-balanced sampling from the full training set. \textbf{RL}: number of samples retained after embedding-variance filtering. \textbf{Retain (\%)}: retention rate. Datasets are grouped by modality.}
	\vspace{2mm}
	\centering
	\small
	\setlength{\tabcolsep}{10pt}
	\renewcommand{\arraystretch}{1.05}
	\begin{tabular}{l r r r}
		\toprule
		\textbf{Dataset} & \textbf{Pre-sample} & \textbf{RL} & \textbf{Retain (\%)} \\
		\midrule
		\rowcolor[HTML]{EDEDED}
		\multicolumn{4}{c}{\emph{Image (14 datasets)}} \\
		ImageNet-1K       &    476 &    287 & 60.3 \\
		N24News           &     24 &     17 & 70.8 \\
		VOC2007           &     20 &      7 & 35.0 \\
		SUN397            &    397 &    148 & 37.3 \\
		OK-VQA            &    575 &    348 & 60.5 \\
		A-OKVQA           &    575 &    364 & 63.3 \\
		DocVQA            &    575 &    125 & 21.7 \\
		InfographicsVQA   &    575 &    323 & 56.2 \\
		ChartQA           &    575 &    298 & 51.8 \\
		Visual7W          &    575 &    267 & 46.4 \\
		VisDial           &    575 &    200 & 34.8 \\
		MSCOCO-i2t        &    575 &    222 & 38.6 \\
		VisualNews-i2t    &    575 &    337 & 58.6 \\
		WebQA             &    574 &    391 & 68.1 \\
		\cmidrule{1-4}
		\emph{Subtotal}   & \emph{6,666} & \emph{3,334} & \emph{50.0} \\
		\midrule
		\rowcolor[HTML]{EDEDED}
		\multicolumn{4}{c}{\emph{Visual Document (3 datasets)}} \\
		ViDoRe            &  2,222 &  1,111 & 50.0 \\
		VisRAG-in-domain  &  2,222 &  1,086 & 48.9 \\
		VisRAG-synthetic  &  2,222 &  1,136 & 51.1 \\
		\cmidrule{1-4}
		\emph{Subtotal}   & \emph{6,666} & \emph{3,333} & \emph{50.0} \\
		\midrule
		\rowcolor[HTML]{EDEDED}
		\multicolumn{4}{c}{\emph{Video (2 datasets)}} \\
		LLaVA-Hound-VQA     &  3,333 &  1,667 & 50.0 \\
		LLaVA-Hound-Caption &  3,333 &  1,667 & 50.0 \\
		\cmidrule{1-4}
		\emph{Subtotal}   & \emph{6,666} & \emph{3,334} & \emph{50.0} \\
		\midrule
		\textbf{Total}    & \textbf{19,998} & \textbf{10,001} & \textbf{50.0} \\
		\bottomrule
	\end{tabular}
	\label{tab:rl_data}
\end{table}

\section{Detailed Scores on MMEB-V2}
\label{sec:detailed_scores_v2}

\definecolor{avgcolor}{RGB}{240,248,255}    
\definecolor{icatcolor}{RGB}{255,245,235}   
\definecolor{vcatcolor}{RGB}{240,255,240}   
\definecolor{vdcatcolor}{RGB}{245,240,255}  

\begin{table}[H]
	\centering
	\caption{Per-dataset scores on the full MMEB-V2 benchmark (78 tasks). Numbers in parentheses represent the task count for each category.}
	\vspace{2mm}
	\renewcommand{\arraystretch}{1.1}
	\resizebox{\textwidth}{!}{
		\begin{tabular}{l|ccccccccccc}
			\toprule
			~                                              & ColPali v1.3 & GME-7B & VLM2Vec-7B & VLM2Vec-V2.0 & CAFe-7B & UME-R1-2B & UME-R1-7B & TTE$_s$-2B & TTE$_s$-7B & TWN-4B & TWN-8B \\
			\midrule
			\rowcolor{avgcolor}
			Avg - All (78 tasks)                           & 44.4         & 57.8   & 52.3       & 58.0         & 60.6    & 60.1      & 64.5      & 63.1        & 68.6        & 66.6      & 68.7      \\
			\midrule
			\rowcolor{avgcolor}
			Avg - Image (36 tasks, Hit@1)                  & 34.9         & 56.0   & 65.5       & 64.9         & 67.6    & 66.6      & 71.3      & 70.1        & 74.2        & 71.3      & 73.4      \\
			\rowcolor{avgcolor}
			Avg - Video (18 tasks, Hit@1)                  & 28.2         & 38.4   & 33.7       & 34.6         & 42.4    & 42.2      & 47.5      & 41.3        & 46.8        & 45.7      & 48.2      \\
			\rowcolor{avgcolor}
			Avg - Visdoc (24 tasks, NDCG@5)                & 71.0         & 75.2   & 46.4       & 65.4         & 63.9    & 63.9      & 67.1      & 68.8        & 76.4        & 75.3      & 77.0      \\
			\midrule
			\rowcolor{icatcolor}
			I-CLS (10)                                     & 40.3         & 57.7   & 62.7       & 62.9         & 63.6    & 64.8      & 67.1      & 67.9        & 69.7        & 68.6      & 70.2      \\
			\rowcolor{icatcolor}
			I-QA (10)                                      & 11.5         & 34.7   & 56.9       & 56.3         & 61.7    & 62.8      & 69.2      & 66.6        & 72.4        & 71.7      & 74.3      \\
			\rowcolor{icatcolor}
			I-RET (12)                                     & 48.1         & 71.2   & 69.4       & 69.5         & 69.1    & 67.6      & 71.9      & 70.2        & 74.0        & 68.0      & 70.8      \\
			\rowcolor{icatcolor}
			I-VG (4)                                       & 40.3         & 59.3   & 82.2       & 77.3         & 87.6    & 77.2      & 84.9      & 84.1        & 90.6        & 87.0      & 87.1      \\
			\rowcolor{vcatcolor}
			V-CLS (5)                                      & 26.7         & 37.4   & 39.1       & 39.3         & 35.8    & 44.3      & 48.6      & 47.3        & 49.1        & 45.8      & 50.1      \\
			\rowcolor{vcatcolor}
			V-QA (5)                                       & 37.8         & 50.4   & 30.0       & 34.3         & 58.7    & 51.0      & 60.7      & 49.1        & 60.6        & 63.4      & 64.0      \\
			\rowcolor{vcatcolor}
			V-RET (5)                                      & 21.6         & 28.4   & 29.0       & 28.8         & 34.4    & 32.9      & 38.2      & 33.2        & 36.4        & 33.6      & 34.8      \\
			\rowcolor{vcatcolor}
			V-MR (3)                                       & 25.5         & 37.0   & 38.9       & 36.8         & 39.5    & 39.7      & 39.3      & 32.1        & 37.2        & 36.6      & 40.8      \\
			\rowcolor{vdcatcolor}
			VD-Vidore-V1 (10)                              & 83.6         & 89.4   & 56.9       & 75.7         & 70.7    & 72.4      & 75.7      & 77.5        & 84.1        & 81.5      & 82.5      \\
			\rowcolor{vdcatcolor}
			VD-Vidore-V2 (4)                               & 52.0         & 55.6   & 9.4        & 45.1         & 49.6    & 46.2      & 50.5      & 53.2        & 62.7        & 53.6      & 56.7      \\
			\rowcolor{vdcatcolor}
			VD-VisRAG (6)                                  & 81.1         & 85.0   & 59.1       & 79.6         & 79.5    & 79.2      & 83.7      & 83.2        & 91.9        & 84.8      & 86.2      \\
			\rowcolor{vdcatcolor}
			VD-OOD (4)                                     & 43.1         & 44.4   & 38.1       & 39.6         & 38.1    & 37.2      & 37.6      & 41.1        & 47.6        & 67.3      & 70.0      \\
			\midrule

			ImageNet-1K                                    & 42.4         & 64.6   & 80.1       & 80.8         & 77.3    & 75.3      & 80.4      & 83.3        & 84.3        & 81.0      & 81.7      \\
			N24News                                        & 25.5         & 50.5   & 79.7       & 72.9         & 83.2    & 81.1      & 82.3      & 78.6        & 83.1        & 75.3      & 75.6      \\
			HatefulMemes                                   & 50.6         & 53.6   & 69.7       & 56.3         & 78.7    & 75.2      & 79.0      & 64.0        & 67.4        & 72.0      & 74.0      \\
			VOC2007                                        & 69.8         & 80.3   & 80.7       & 85.0         & 89.8    & 80.0      & 90.8      & 86.3        & 86.6        & 86.0      & 89.4      \\
			SUN397                                         & 56.1         & 69.5   & 77.4       & 71.0         & 79.9    & 79.4      & 80.3      & 77.5        & 78.9        & 80.9      & 81.4      \\
			Place365                                       & 27.5         & 39.1   & 37.4       & 35.9         & 45.0    & 42.6      & 46.8      & 45.7        & 44.6        & 45.1      & 47.2      \\
			ImageNet-A                                     & 14.9         & 41.2   & 58.1       & 47.4         & 55.2    & 50.4      & 53.9      & 50.9        & 60.4        & 59.8      & 59.9      \\
			ImageNet-R                                     & 64.6         & 83.9   & 73.9       & 89.3         & 88.0    & 88.7      & 90.1      & 89.7        & 90.5        & 90.1      & 91.7      \\
			ObjectNet                                      & 45.6         & 69.0   & 40.1       & 65.2         & 22.5    & 52.0      & 42.3      & 74.1        & 72.6        & 76.6      & 77.4      \\
			Country211                                     & 6.0          & 24.8   & 29.8       & 25.2         & 16.7    & 23.4      & 25.0      & 28.5        & 29.0        & 18.9      & 23.7      \\
			OK-VQA                                         & 9.4          & 33.2   & 56.8       & 51.5         & 67.3    & 62.4      & 71.7      & 68.4        & 74.7        & 70.6      & 71.1      \\
			A-OKVQA                                        & 6.6          & 21.0   & 47.3       & 43.6         & 63.8    & 51.1      & 58.7      & 57.1        & 66.1        & 62.6      & 63.9      \\
			DocVQA                                         & 11.3         & 41.4   & 89.7       & 90.1         & 79.2    & 92.2      & 93.8      & 94.2        & 95.6        & 94.7      & 94.7      \\
			InfographicsVQA                                & 5.0          & 20.3   & 60.0       & 58.8         & 53.3    & 67.7      & 79.2      & 65.6        & 77.5        & 80.0      & 82.3      \\
			ChartQA                                        & 5.7          & 17.8   & 56.9       & 47.4         & 48.8    & 64.9      & 75.1      & 57.5        & 70.9        & 81.7      & 84.4      \\
			Visual7W                                       & 6.1          & 22.2   & 52.7       & 52.9         & 52.5    & 54.1      & 55.2      & 54.1        & 57.9        & 57.8      & 62.9      \\
			ScienceQA                                      & 16.3         & 28.0   & 38.5       & 38.2         & 65.4    & 42.7      & 53.7      & 50.7        & 60.0        & 56.5      & 61.2      \\
			VizWiz                                         & 27.6         & 39.0   & 39.9       & 43.3         & 43.8    & 46.8      & 51.6      & 55.1        & 53.8        & 55.2      & 57.8      \\
			GQA                                            & 8.3          & 76.9   & 55.1       & 64.9         & 65.7    & 67.3      & 69.3      & 77.0        & 80.9        & 75.4      & 78.0      \\
			TextVQA                                        & 18.8         & 46.8   & 71.6       & 72.2         & 76.8    & 78.6      & 83.5      & 86.2        & 87.0        & 82.4      & 86.7      \\
			VisDial                                        & 41.2         & 60.8   & 81.9       & 82.7         & 82.7    & 76.6      & 80.7      & 81.2        & 84.4        & 80.8      & 84.0      \\
			CIRR                                           & 8.2          & 54.9   & 51.1       & 57.5         & 60.4    & 53.7      & 55.3      & 59.4        & 65.1        & 47.5      & 47.9      \\
			VisualNews\_t2i                                & 50.1         & 79.7   & 80.5       & 74.5         & 69.5    & 71.7      & 76.8      & 72.8        & 78.5        & 72.0      & 75.1      \\
			VisualNews\_i2t                                & 47.6         & 83.6   & 81.2       & 78.2         & 79.4    & 74.2      & 82.0      & 76.5        & 81.3        & 77.3      & 79.8      \\
			MSCOCO\_t2i                                    & 59.2         & 71.2   & 77.2       & 75.3         & 75.4    & 75.1      & 78.3      & 75.2        & 77.9        & 77.6      & 79.2      \\
			MSCOCO\_i2t                                    & 49.9         & 57.7   & 73.9       & 71.4         & 73.1    & 68.9      & 71.4      & 71.1        & 73.1        & 75.6      & 77.3      \\
			NIGHTS                                         & 65.5         & 67.6   & 67.6       & 68.6         & 66.7    & 67.2      & 68.1      & 70.8        & 69.8        & 67.4      & 71.4      \\
			WebQA                                          & 53.8         & 91.4   & 88.3       & 90.6         & 89.3    & 90.0      & 90.9      & 90.4        & 90.8        & 88.5      & 90.3      \\
			FashionIQ                                      & 5.9          & 37.8   & 17.1       & 19.5         & 39.0    & 17.1      & 23.4      & 26.3        & 29.7        & 9.5      & 18.5      \\
			Wiki-SS-NQ                                     & 80.5         & 78.2   & 62.3       & 66.9         & 61.2    & 62.0      & 72.5      & 64.2        & 70.5        & 70.9      & 72.0      \\
			OVEN                                           & 50.0         & 75.1   & 66.5       & 64.3         & 60.8    & 66.9      & 71.4      & 67.6        & 72.7        & 63.7      & 66.8      \\
			EDIS                                           & 64.7         & 96.0   & 85.7       & 84.1         & 71.3    & 88.0      & 92.0      & 87.0        & 93.9        & 85.8      & 87.3      \\
			MSCOCO                                         & 36.7         & 31.4   & 75.7       & 67.1         & 84.7    & 69.5      & 72.7      & 67.7        & 74.1        & 78.2      & 78.1      \\
			RefCOCO                                        & 64.5         & 60.9   & 87.6       & 87.1         & 89.4    & 83.3      & 91.4      & 91.4        & 97.7        & 92.5      & 93.8      \\
			RefCOCO-Matching                               & 3.9          & 78.4   & 84.6       & 85.8         & 83.0    & 84.4      & 91.1      & 95.0        & 96.3        & 91.1      & 91.5      \\
			Visual7W-Pointing                              & 56.1         & 66.5   & 81.0       & 69.2         & 93.2    & 71.5      & 84.2      & 82.5        & 94.3        & 86.0      & 85.0      \\
			\midrule
			K700                                           & 23.4         & 39.7   & 35.5       & 38.0         & 40.1    & 35.8      & 42.8      & 49.6        & 55.0        & 52.2      & 55.4      \\
			SmthSmthV2                                     & 25.1         & 30.6   & 32.1       & 42.8         & 35.8    & 44.1      & 50.4      & 50.4        & 44.9        & 33.0      & 36.8      \\
			HMDB51                                         & 24.8         & 47.9   & 42.2       & 40.9         & 46.9    & 54.4      & 58.3      & 52.5        & 51.7        & 45.8      & 50.7      \\
			UCF101                                         & 49.4         & 54.7   & 61.8       & 60.0         & 39.6    & 67.2      & 70.0      & 58.3        & 64.2        & 70.9      & 76.1      \\
			Breakfast                                      & 10.9         & 14.3   & 23.8       & 14.8         & 16.6    & 20.1      & 21.5      & 25.4        & 29.7        & 27.0      & 31.5      \\
			MVBench                                        & 33.7         & 46.6   & 28.5       & 33.7         & 48.9    & 49.9      & 58.2      & 48.5        & 59.5        & 62.6      & 62.6      \\
			Video-MME                                      & 30.6         & 39.2   & 27.8       & 30.7         & 46.0    & 41.7      & 47.3      & 45.8        & 53.1        & 55.0      & 55.3      \\
			NExTQA                                         & 35.2         & 53.6   & 20.3       & 20.9         & 62.4    & 59.9      & 69.6      & 53.8        & 70.1        & 67.1      & 67.4      \\
			EgoSchema                                      & 38.4         & 46.8   & 21.8       & 34.0         & 60.0    & 45.4      & 52.4      & 36.4        & 55.6        & 54.2      & 54.6      \\
			ActivityNetQA                                  & 51.3         & 65.6   & 51.4       & 52.3         & 76.0    & 57.8      & 76.0      & 60.8        & 64.6        & 77.9      & 80.1      \\
			DiDeMo                                         & 22.8         & 26.4   & 29.3       & 30.4         & 37.8    & 32.4      & 40.0      & 33.5        & 34.9        & 31.8      & 33.5      \\
			MSR-VTT                                        & 17.6         & 31.8   & 34.5       & 28.3         & 36.5    & 34.3      & 38.9      & 34.8        & 37.6        & 37.1      & 38.2      \\
			MSVD                                           & 45.4         & 49.7   & 46.7       & 48.1         & 56.4    & 55.4      & 60.8      & 56.5        & 58.5        & 54.8      & 56.5      \\
			VATEX                                          & 16.7         & 24.9   & 25.5       & 26.5         & 32.0    & 29.9      & 32.6      & 25.6        & 31.0        & 29.0      & 29.8      \\
			YouCook2                                       & 5.3          & 9.1    & 9.0        & 10.6         & 9.5     & 12.7      & 18.5      & 15.8        & 19.9        & 15.3      & 16.0      \\
			QVHighlight                                    & 19.9         & 59.5   & 57.7       & 49.4         & 58.4    & 57.5      & 54.9      & 38.9        & 51.0        & 45.0      & 50.2      \\
			Charades-STA                                   & 29.0         & 14.0   & 19.8       & 20.2         & 18.7    & 20.4      & 21.9      & 19.5        & 18.9        & 23.2      & 25.3      \\
			MomentSeeker                                   & 27.6         & 37.4   & 39.3       & 40.8         & 41.4    & 41.2      & 41.1      & 37.7        & 41.5        & 41.6      & 46.9      \\
			\midrule
			ViDoRe\_arxivqa                                & 81.7         & 86.9   & 60.2       & 80.6         & 73.3    & 73.9      & 73.6      & 80.7        & 84.6        & 88.4      & 89.1      \\
			ViDoRe\_docvqa                                 & 56.6         & 57.5   & 34.7       & 44.9         & 38.3    & 37.9      & 41.1      & 44.5        & 46.0        & 48.5      & 50.2      \\
			ViDoRe\_infovqa                                & 84.9         & 91.6   & 70.4       & 83.7         & 80.6    & 76.2      & 80.8      & 84.8        & 88.7        & 88.2      & 89.8      \\
			ViDoRe\_tabfquad                               & 86.9         & 94.6   & 78.2       & 89.2         & 80.7    & 86.1      & 90.2      & 88.4        & 94.7        & 91.2      & 92.6      \\
			ViDoRe\_tatdqa                                 & 70.9         & 74.1   & 27.6       & 43.8         & 37.8    & 40.6      & 46.7      & 50.4        & 59.4        & 52.7      & 54.4      \\
			ViDoRe\_shiftproject                           & 75.1         & 96.8   & 38.6       & 60.8         & 52.0    & 66.8      & 65.0      & 65.2        & 81.6        & 75.1      & 76.3      \\
			ViDoRe\_artificial\_intelligence               & 95.7         & 99.6   & 67.7       & 88.5         & 86.0    & 85.9      & 89.5      & 91.9        & 98.1        & 97.4      & 97.2      \\
			ViDoRe\_energy                                 & 94.7         & 95.3   & 60.4       & 86.5         & 84.8    & 83.3      & 85.7      & 88.7        & 93.5        & 88.8      & 89.4      \\
			ViDoRe\_government\_reports                    & 93.6         & 98.8   & 61.8       & 85.0         & 85.0    & 82.6      & 89.8      & 86.9        & 96.7        & 89.2      & 90.5      \\
			ViDoRe\_healthcare\_industry                   & 95.9         & 99.3   & 69.9       & 92.2         & 88.4    & 90.8      & 94.3      & 92.8        & 97.9        & 95.6      & 95.5      \\
			ViDoRe\_esg\_reports\_human\_labeled\_v2       & 51.3         & 63.4   & 6.8        & 45.6         & 50.7    & 50.2      & 50.4      & 59.0        & 69.4        & 54.8      & 55.3      \\
			ViDoRe\_biomedical\_lectures\_v2\_multilingual & 54.7         & 49.5   & 5.1        & 44.3         & 50.9    & 46.2      & 50.7      & 52.0        & 60.8        & 54.1      & 59.7      \\
			ViDoRe\_economics\_reports\_v2\_multilingual   & 49.0         & 54.2   & 13.9       & 43.0         & 54.3    & 45.7      & 57.8      & 49.8        & 60.4        & 57.9      & 59.7      \\
			ViDoRe\_esg\_reports\_v2\_multilingual         & 52.9         & 55.4   & 11.9       & 46.6         & 42.3    & 42.7      & 43.2      & 52.1        & 60.3        & 47.6      & 52.1      \\
			VisRAG\_ArxivQA                                & 80.9         & 87.4   & 52.6       & 76.9         & 74.0    & 74.3      & 80.5      & 78.5        & 94.5        & 85.4      & 87.0      \\
			VisRAG\_ChartQA                                & 72.3         & 86.1   & 57.7       & 83.7         & 82.7    & 86.0      & 85.0      & 84.4        & 91.2        & 84.0      & 84.0      \\
			VisRAG\_MP-DocVQA                              & 82.0         & 89.7   & 60.6       & 88.1         & 75.1    & 75.6      & 83.4      & 79.2        & 90.1        & 84.4      & 87.3      \\
			VisRAG\_SlideVQA                               & 85.1         & 92.6   & 54.7       & 84.1         & 87.6    & 87.1      & 91.5      & 92.3        & 95.6        & 92.6      & 94.1      \\
			VisRAG\_InfoVQA                                & 83.5         & 88.6   & 66.0       & 82.3         & 87.9    & 84.4      & 89.2      & 87.2        & 93.0        & 90.8      & 91.9      \\
			VisRAG\_PlotQA                                 & 79.3         & 76.5   & 62.7       & 75.9         & 69.4    & 68.0      & 72.7      & 77.5        & 86.9        & 71.5      & 72.9      \\
			ViDoSeek-page                                  & 38.1         & 32.6   & 16.3       & 29.1         & 22.5    & 21.2      & 21.3      & 22.6        & 35.0        & 81.3      & 86.6      \\
			ViDoSeek-doc                                   & 87.5         & 90.3   & 69.4       & 79.0         & 73.8    & 75.9      & 75.3      & 82.0        & 84.4        & 82.8      & 84.6      \\
			MMLongBench-page                               & 27.1         & 36.9   & 0.4        & 15.8         & 13.3    & 11.9      & 12.3      & 12.9        & 20.7        & 53.4      & 55.5      \\
			MMLongBench-doc                                & 80.4         & 85.2   & 28.8       & 63.0         & 42.6    & 39.7      & 41.3      & 47.0        & 50.4        & 51.8      & 53.3      \\
			\bottomrule
		\end{tabular}
	}
	\label{tab:detailed_v2}
\end{table}
\clearpage

\section{Training Dynamics}
\label{sec:training_dynamics}

We visualize key training metrics to provide insight into the optimization behavior across stages.

\begin{figure}[!ht]
	\centering
	\includegraphics[width=\textwidth]{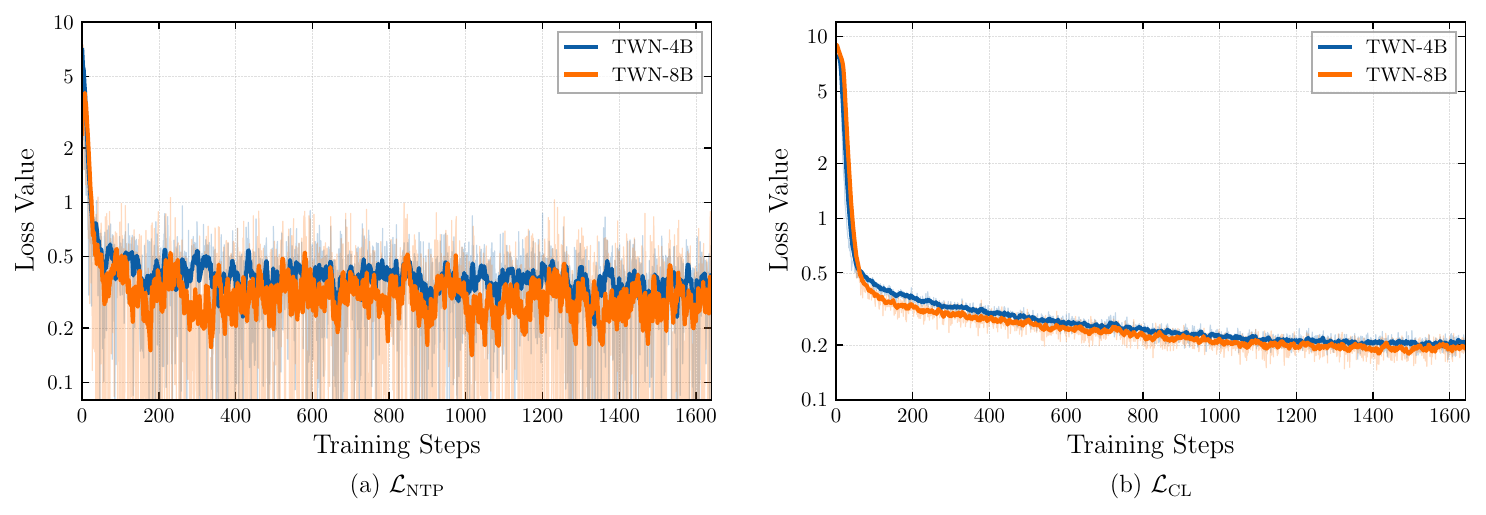}
	\caption{Training loss curves during Stage~1 (SFT) for TWN-4B and TWN-8B. (a)~Next-token prediction loss $\mathcal{L}_{\text{NTP}}$ for the reasoning adapter. (b)~Average contrastive loss $\mathcal{L}_{\text{CL}}$ for the embedding adapter. Faint lines show per-step values; bold lines show exponential moving averages. Both axes use logarithmic scale.}
	\label{fig:sft_loss}
\end{figure}

\paragraph{SFT Stage.}
Figure~\ref{fig:sft_loss} shows the training loss curves for both TWN-4B and TWN-8B during Stage~1 (SFT).
The next-token prediction loss $\mathcal{L}_{\text{NTP}}$ (Figure~\ref{fig:sft_loss}a) drops sharply within the first ${\sim}100$ steps and stabilizes around $0.3$--$0.4$, indicating that both models quickly learn to generate structured CoT traces.
TWN-8B converges to a slightly lower NTP loss than TWN-4B, consistent with its larger model capacity.
The contrastive loss $\mathcal{L}_{\text{CL}}$ (Figure~\ref{fig:sft_loss}b) exhibits a smooth, monotonic decrease from ${\sim}9$ to ${\sim}0.2$, reflecting steady improvement in embedding discriminability throughout training.
Both models follow nearly identical contrastive loss trajectories.
Notably, both losses decrease smoothly without oscillation or divergence, consistent with the hypothesis that detaching gradients between the reasoning and embedding adapters helps mitigate gradient conflict, contributing to stable joint optimization.

\begin{figure}[!ht]
	\centering
	\includegraphics[width=\textwidth]{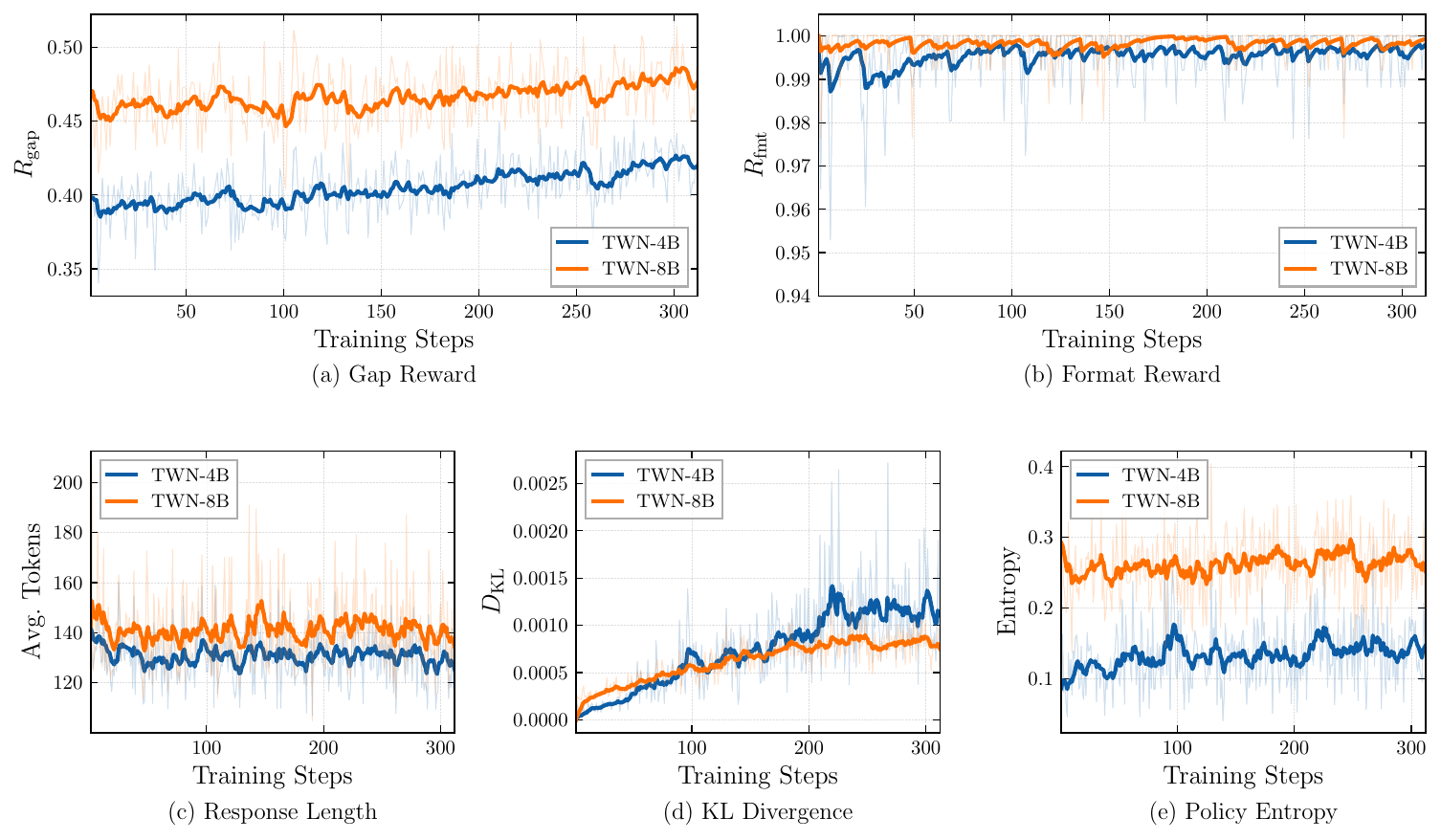}
	\caption{Training dynamics during Stage~2 (RL) for TWN-4B and TWN-8B. (a)~Gap reward $R_{\text{gap}}$. (b)~Format reward $R_{\text{fmt}}$. (c)~Average response length in tokens. (d)~KL divergence from the reference policy. (e)~Policy entropy. Faint lines show per-step values; bold lines show exponential moving averages.}
	\label{fig:rl_dynamics}
\end{figure}

\paragraph{RL Stage.}
Figure~\ref{fig:rl_dynamics} tracks five key metrics during Stage~2 (RL) for both TWN-4B and TWN-8B.
The gap reward $R_{\text{gap}}$ (Figure~\ref{fig:rl_dynamics}a) increases steadily throughout training, indicating that the RL-optimized CoT produces embeddings with better positive--negative separation than the SFT initialization.
TWN-8B maintains a consistently higher gap reward, reflecting its stronger reasoning capacity.
The format reward $R_{\text{fmt}}$ (Figure~\ref{fig:rl_dynamics}b) rapidly saturates near $1.0$, indicating that both models reliably produce well-structured \texttt{<think>...<\/think><answer>...<\/answer>} outputs.
The average response length (Figure~\ref{fig:rl_dynamics}c) remains stable around $130$--$150$ tokens without exhibiting the reward hacking behavior (unbounded length growth) sometimes observed in RL for language generation.
KL divergence from the reference policy (Figure~\ref{fig:rl_dynamics}d) grows gradually and remains small ($<0.003$), indicating that the policy explores beyond the SFT distribution but does not diverge excessively.
Policy entropy (Figure~\ref{fig:rl_dynamics}e) increases moderately from ${\sim}0.1$ to ${\sim}0.2$--$0.3$, reflecting healthy exploration: the policy diversifies its reasoning strategies rather than collapsing to a single mode.

\clearpage
\section{Case Studies}
\label{sec:case_studies}

We present qualitative examples to illustrate when chain-of-thought reasoning improves retrieval and when it is unnecessary or even harmful.
Each case shows the query, target, generated CoT trace, and retrieval results under three inference modes: base ($w{=}0$), cot ($w{=}1$), and adaptive (routing gate decides).

\subsection{When CoT Helps}

\begin{figure}[!htb]
	\centering
	\includegraphics[width=\textwidth]{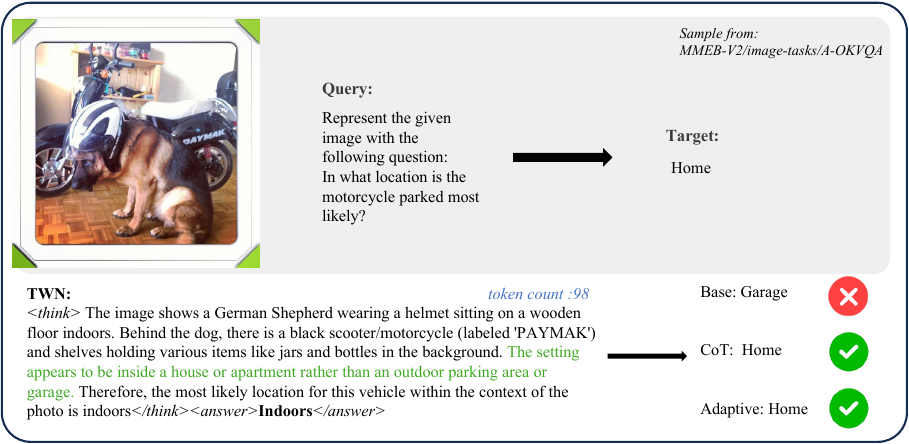}
	\caption{Positive case 1 (A-OKVQA): CoT reasons about the indoor domestic setting to correctly retrieve ``Home'' instead of the superficial association ``Garage.''}
	\label{fig:cot_pos_1}
\end{figure}

\begin{figure}[!htb]
	\centering
	\includegraphics[width=\textwidth]{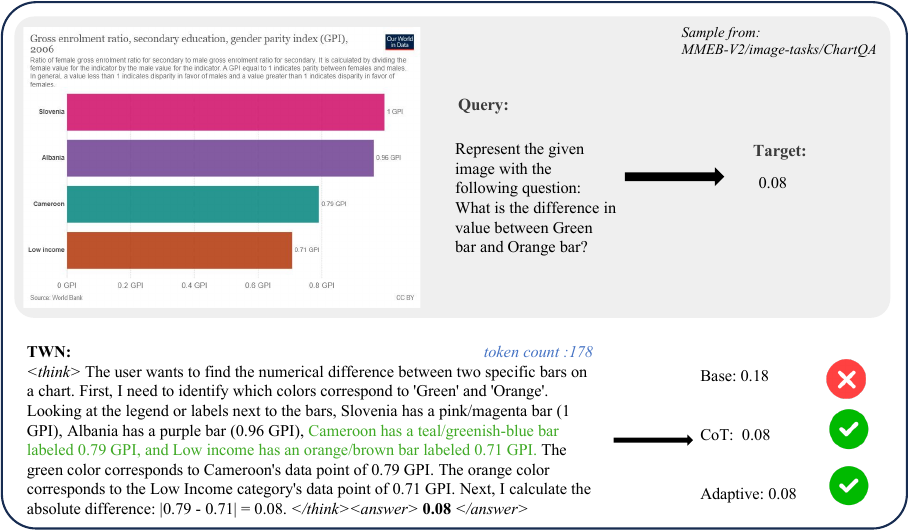}
	\caption{Positive case 2 (ChartQA): CoT reads specific bar values and computes the numerical difference, which requires multi-step reasoning beyond visual similarity matching.}
	\label{fig:cot_pos_2}
\end{figure}

\begin{figure}[!htb]
	\centering
	\includegraphics[width=\textwidth]{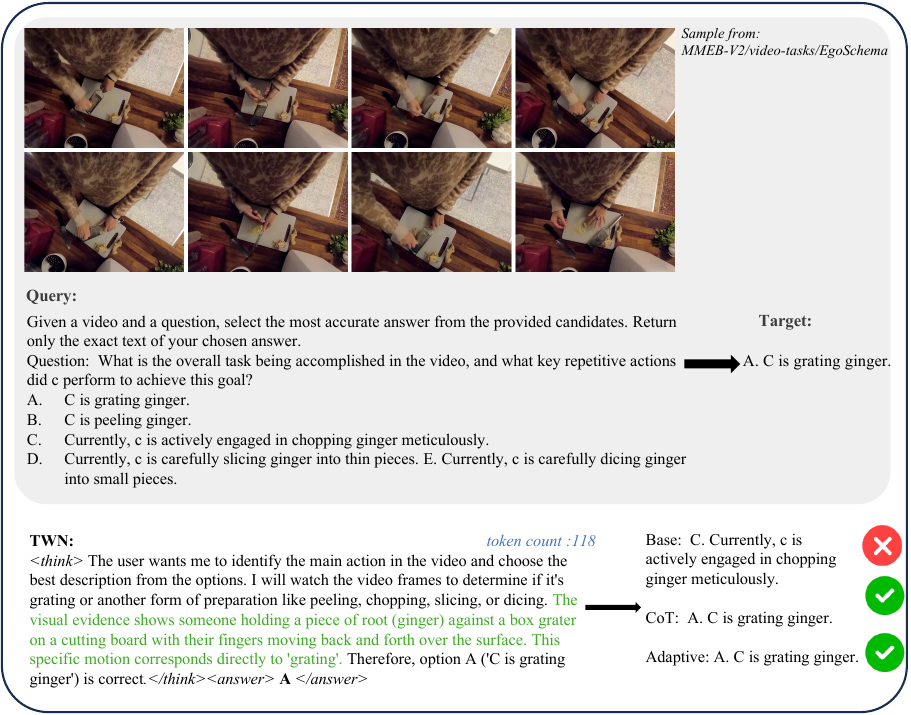}
	\caption{Positive case 3 (EgoSchema): CoT analyzes the hand motion pattern across video frames to distinguish ``grating'' from ``chopping,'' a fine-grained action recognition task.}
	\label{fig:cot_pos_3}
\end{figure}

Figures~\ref{fig:cot_pos_1}--\ref{fig:cot_pos_3} show cases where the base mode retrieves incorrectly but the cot mode succeeds, and the adaptive routing gate correctly triggers reasoning.
In Figure~\ref{fig:cot_pos_1}, the query asks about the likely parking location of a motorcycle in a domestic scene. The base embedding superficially associates ``motorcycle'' with ``Garage,'' while CoT reasons about the indoor setting (wooden floor, shelves, dog) to correctly infer ``Home.''
In Figure~\ref{fig:cot_pos_2}, the query requires reading specific bar values from a chart and computing their numerical difference. The base mode retrieves a wrong value (0.18), while CoT identifies each bar's label and computes $|0.79 - 0.71| = 0.08$ correctly.
In Figure~\ref{fig:cot_pos_3}, the query involves a video understanding task that requires distinguishing between fine-grained actions (grating vs.\ chopping). CoT analyzes the hand motion against a box grater across frames to correctly select ``grating.''

\FloatBarrier
\subsection{When CoT Is Unnecessary}

\begin{figure}[!htb]
	\centering
	\includegraphics[width=\textwidth]{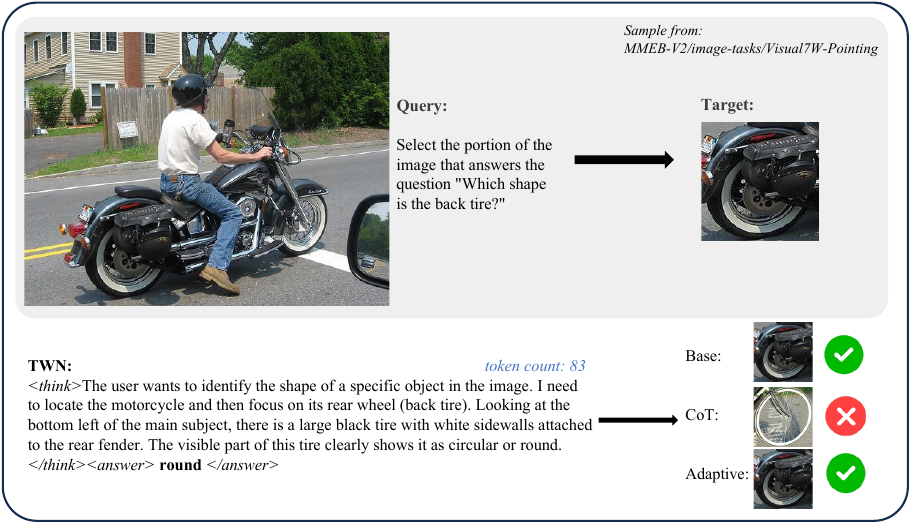}
	\caption{Negative case 1 (Visual7W-Pointing): A simple visual grounding task where CoT's unnecessary reasoning leads to retrieving the wrong image crop, while base mode succeeds directly.}
	\label{fig:cot_neg_1}
\end{figure}

\begin{figure}[!htb]
	\centering
	\includegraphics[width=\textwidth]{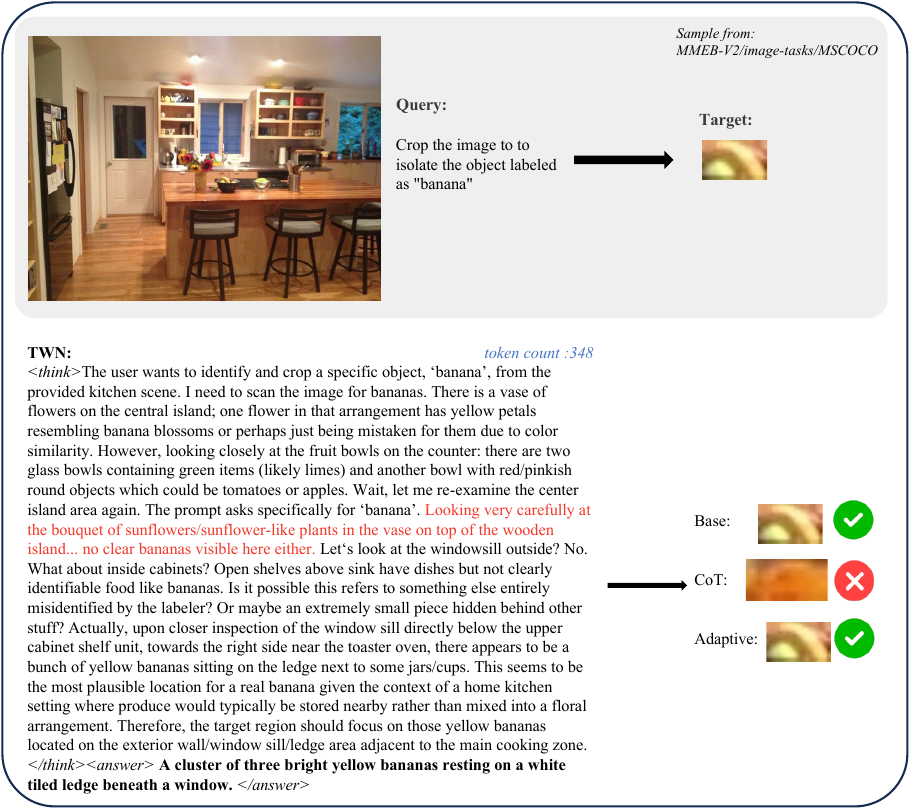}
	\caption{Negative case 2 (MSCOCO): CoT generates 348 tokens of overthinking that confuses visual elements, while base mode correctly localizes the target object with zero reasoning overhead.}
	\label{fig:cot_neg_2}
\end{figure}

\begin{figure}[!htb]
	\centering
	\includegraphics[width=\textwidth]{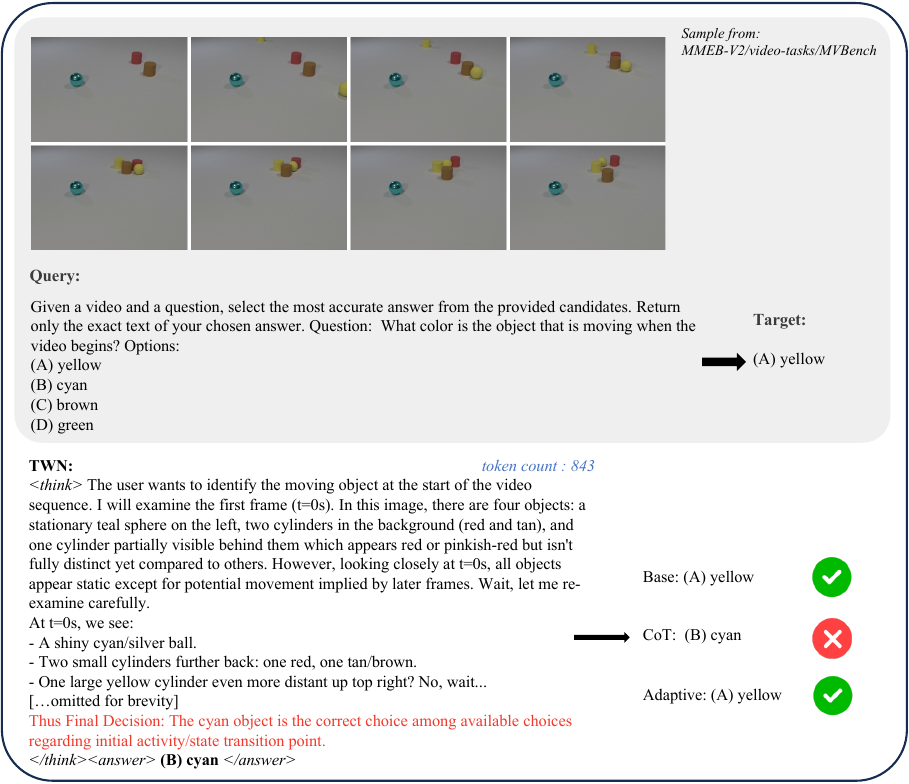}
	\caption{Negative case 3 (MVBench): An 843-token CoT trace with extensive self-correction ultimately hallucinates the wrong answer, while base mode retrieves correctly through direct visual matching.}
	\label{fig:cot_neg_3}
\end{figure}

Figures~\ref{fig:cot_neg_1}--\ref{fig:cot_neg_3} show cases where CoT reasoning is unnecessary or even harmful, and the adaptive routing gate correctly avoids or overrides it.
In Figure~\ref{fig:cot_neg_1}, the query asks about the shape of a motorcycle's back tire---a visually straightforward grounding task. The base embedding directly matches the correct crop, while CoT's redundant reasoning about ``circular or round'' leads to retrieving the wrong image region. The adaptive mode correctly selects base.
In Figure~\ref{fig:cot_neg_2}, the query asks to crop a banana from a kitchen scene. The base embedding localizes the banana directly, but CoT generates an excessively long trace (348 tokens) that overthinks the scene, confusing sunflowers with bananas and ultimately retrieving an incorrect crop. The adaptive mode avoids this failure.
In Figure~\ref{fig:cot_neg_3}, a video QA task asks about the color of a moving object. The base embedding correctly identifies ``yellow,'' but CoT produces an extremely long trace (843 tokens) with extensive self-correction that ultimately arrives at the wrong answer ``cyan.'' This illustrates that excessive reasoning can introduce hallucination on tasks where direct visual matching suffices.

\end{document}